%% file: main.tex
\documentclass[11pt]{article}

\usepackage[preprint]{acl}

\usepackage{times}
\usepackage{latexsym}
\usepackage[T1]{fontenc}
\usepackage[utf8]{inputenc}
\usepackage{microtype}
\usepackage{inconsolata}
\usepackage{graphicx}
\usepackage{booktabs}
\usepackage{enumitem}
\usepackage{listings}
\usepackage{amsmath}
\usepackage{scalefnt}
\usepackage{colortbl}
\usepackage{longtable}

\graphicspath{{figures/}{./}}

\title{PlotTwist: A Creative Plot Generation Framework with Small Language Models}

\author{
  \begin{tabular}[t]{@{}c@{\hspace{2.5em}}c@{\hspace{2.5em}}c@{}}
    \textbf{Abhinav Thorat}\thanks{\ Equal contribution.} & \textbf{Ravi Kolla}\footnotemark[1] & \textbf{Jyotin Goel}\thanks{\ Work done during an internship at Sony Research India.} \\[2pt]
    \mdseries\small Sony Research India & \mdseries\small Sony Research India & \mdseries\small Sony Research India \\[1pt]
    \mdseries\small\texttt{abhinav.thorat@sony.com} & \mdseries\small\texttt{ravi.kolla@sony.com} & \mdseries\small\texttt{jyotin.goel@sony.com} \\[10pt]
    \multicolumn{3}{c}{%
      \begin{tabular}[t]{@{}c@{\hspace{2.5em}}c@{}}
        \textbf{Madhav Kataria}\footnotemark[2] & \textbf{Niranjan Pedanekar} \\[2pt]
        \mdseries\small Sony Research India & \mdseries\small Sony Research India \\[1pt]
        \mdseries\small\texttt{kataria.madhav@sony.com} & \mdseries\small\texttt{niranjan.pedanekar@sony.com} \\
      \end{tabular}%
    } \\
  \end{tabular}
}

\begin{document}
\maketitle
\input{sections/0_abstract}

\input{sections/1_intro}

\input{sections/2_related}
\input{sections/3_problem}

\input{sections/4_framework}

\input{sections/5_experiments}

\input{sections/5b_jury}
\input{sections/6_conclusion}

\clearpage
\input{sections/7_limitations}

\bibliography{references}

\onecolumn
\input{sections/A_appendix}

\end{document}

%% file: sections/0_abstract.tex
\begin{abstract}
\scalefont{0.97}
Creative plot generation presents a fundamental challenge for language models:  transforming a concise premise into a coherent narrative while satisfying multiple narrative constraints, including global coherence, character development, pacing, tone consistency, and emotional progression. Although recent Large Language Models (LLMs) demonstrate
strong fluency on general-purpose tasks, they require preference
alignment to perform well on domain-specific tasks such as creative
plot generation. However, conducting such alignment at the scale of frontier
LLMs is computationally prohibitive, significantly limiting
accessibility and practical deployment. To address this, we present
\textsc{PlotTwist}, a structured framework that enables Small Language
Models (SLMs) with $\leq 3$B active parameters to generate
high-quality, premise-conditioned plots competitive with frontier
systems of vastly greater parameter scale. Our approach decomposes generation
into three specialized components: (1) an Aspect Rating Reward Model,
trained via a novel Positive--Negative prompting strategy; (2) a Mixture-of-Experts (MoE) plot generator
aligned via Direct Preference Optimization (DPO); and (3) an Agentic Evaluation module containing a
cross-family jury that emulates human critical judgment for unbiased,
independent post-hoc assessment. Extensive experiments demonstrate that
\textsc{PlotTwist} consistently outperforms all baselines, including
frontier models, across multiple Narrative Quality Dimensions (NQDs), achieving higher win
rates against all but the strongest baseline, with which it remains
competitive. Further validation
confirms strong sensitivity to narrative quality, as the framework
reliably distinguishes plots derived from critically acclaimed versus
widely panned screenplays. Together, these results establish
structured, preference-based alignment as a resource-efficient approach
to high-quality creative plot generation. \href{https://abhinavthorat.github.io/plottwist/}{\underline{Project Page}}.

\end{abstract}

%% file: sections/1_intro.tex
\section{Introduction}
\label{sec:introduction}
Writers across film studios, streaming platforms, and publishing houses constantly face the same challenge: transforming a concise creative premise into a compelling narrative outline under tight deadlines. A showrunner needs to develop episode arcs for a new series. A screenwriter must pitch three distinct treatments by week's end. An educator requires diverse story examples for a creative writing course. In each case, the task is not simply to generate text, but to craft plots that exhibit coherent structure, believable character arcs, consistent tone, and emotionally resonant turning points, qualities that distinguish professional storytelling from arbitrary event sequences~\cite{teleki2025survey,yao2019plan}. While experienced writers navigate these demands through years of
training and intuition, computational systems offer no comparable
mechanism for structured creative assistance, a gap this work
seeks to address.

The challenge of creative plot generation with generative models extends beyond surface-level text production. Unlike summarization or question-answering, where local context often suffices, plot generation demands long-horizon reasoning over concise conditioning signals. A promising premise, such as ``\texttt{a romantic comedy set in the modern tech startup era}'', provides minimal concrete guidance to writers, yet must expand into a causally connected sequence of events spanning setup, development, climax, and resolution. The narrative must maintain global coherence while ensuring that early character motivations align with later decisions, that tonal shifts feel earned rather than arbitrary, and that pacing sustains engagement across the entire arc. These requirements pose significant difficulties for standard autoregressive language models, which optimize token-level likelihoods and lack explicit mechanisms for enforcing discourse-level constraints. 
Prior work has shown that hierarchical planning and explicit structural decomposition can improve narrative consistency~\cite{fan2018hierarchical,gurung2025learning,teleki2025survey,yao2019plan} but such approaches typically assume large model capacities, task-specific supervision, or relaxed efficiency constraints, leaving open the question of whether effective plot generation is achievable under computational constraints.

Recently, LLMs have demonstrated impressive fluency across creative writing tasks, yet their success comes at a steep cost. Frontier models such as GPT-4.1, Claude Sonnet~4, and Gemini 2.0 Flash operate at vastly greater parameter
scales, demanding substantial computational infrastructure for both training and inference, costs that are ultimately passed on to end users. Beyond raw cost, the literature consistently shows that task-specific alignment yields meaningfully better performance than relying on general-purpose models alone~\cite{ouyang2022training,sun2023evaluating}. In creative writing, this alignment imperative is compounded by a
deeper challenge: scale alone does not reliably resolve long-horizon coherence. Even the largest models exhibit narrative drift, inconsistent characterization, and structural incoherence when generating extended plots without additional inductive biases~\cite{fan2018hierarchical,yao2019plan}. Achieving professional-grade plot generation thus requires targeted alignment to the creative domain, yet for models of this scale, such alignment is computationally prohibitive, particularly when the resulting system is intended for a narrow, specialized use case. 

This observation motivates a natural question: can Small Language Models (SLMs), defined here as models with $\leq 3$B active
parameters per token, generate creative plots of comparable quality to frontier LLMs when aligned using an appropriate structural scaffolding? We hypothesize that the key lies not in model scale, but in externalizing narrative structure into explicit evaluative and training signals. Rather than relying on a monolithic model to implicitly learn all aspects of narrative quality through token prediction, we propose decomposing the generation process into specialized, modular components.
This architectural separation enables SLMs to leverage explicit guidance where large models rely on emergent capabilities, effectively trading model capacity for structured workflow design.
To operationalize this approach, we introduce \textsc{PlotTwist}, a three-component framework for concise premise-conditioned plot generation with SLMs. The first component is an \textit{Aspect Rating Reward Model} that evaluates plots across five Narrative Quality Dimensions (NQDs): character development, tone consistency, pacing, narrative coherence, and emotional turning points. The second component is a \textit{Mixture-of-Experts (MoE) Plot Generator} based on Qwen-3-30B-A3B (3B active parameters), trained via Direct Preference Optimization (DPO) \cite{rafailov2023direct} on preference pairs derived from the aspect rating reward model. The third component is an \textit{Agentic Evaluation Module with a Cross-Family Jury} that operates independently of the training pipeline, providing post-hoc assessment through structured, weakness-focused criteria. 
The key contributions of this work are as follows.
\begin{itemize}[leftmargin=*,noitemsep,topsep=2pt]
\item \textbf{Structured Workflow using SLMs for Plot Generation.} We propose a modular framework comprising an Aspect Rating Reward Model, a DPO-trained MoE Plot Generator, and an independent Agentic Evaluation Module.
\item \textbf{Positive--Negative Prompting for Aspect Rating Reward Modeling.} We introduce a novel prompting strategy that mitigates positivity bias in LLM-based evaluation, enabling reliable aspect-level supervision across five NQDs. 
\item \textbf{Agentic Evaluation using a Cross-Family Jury.} We develop a cross-family jury of five open-weight LLMs, together with a pre-registered evaluation protocol incorporating structured deliberation, a held-out judge, and reliability analyses.
\item \textbf{External Validation of Evaluation Components.} We demonstrate that both the reward model and the agentic evaluator reliably distinguish acclaimed from critically panned plots across all NQDs. 
\item \textbf{Efficient, Quality-Adaptive Plot Generation.} We show that \textsc{PlotTwist} outperforms all baselines across multiple NQDs and achieves higher per-premise win rates against all but the strongest baseline, with which it remains competitive, despite using only 3B active parameters. Moreover,
\textsc{PlotTwist} exhibits principled intervention scaling across quality strata, lightly refining strong narratives while substantially restructuring weak ones rather than uniformly inflating scores.
\end{itemize}

%% file: sections/2_related.tex
\section{Related Work}
\label{sec:literature-survey}
\textbf{Story and Plot Generation.}
Early neural methods introduced hierarchical generation, decomposing stories into premise and continuation stages~\cite{fan2018hierarchical}, while Plan-and-Write frameworks explicitly separated outline planning from surface realization~\cite{yao2019plan}. 
More recently, Agents' Room~\cite{huot2024agents} simulates professional writing rooms. Such systems remain tethered to frontier-scale computation, and a recent survey notes that structured, quality-aware generation remains an open problem~\cite{teleki2025survey}. \\
\textbf{Preference Alignment and Efficient Models.}
Reinforcement Learning from Human Feedback (RLHF) is effective for aligning LLMs with human preferences, but the training pipeline can be computationally expensive. Direct Preference Optimization (DPO)~\cite{rafailov2023direct} offers a simpler, more stable alternative, making it attractive for resource-constrained settings, as is sparse computation in MoE architectures~\cite{shazeer2017outrageously,fedus2022switch}. \\
\textbf{Evaluation of Creative Text and Reliability of LLM Judges.}
Evaluating creative generation remains challenging. Despite their strong surface-level fluency, LLM-generated stories often lack authentic creativity~\cite{chakrabarty2024art}. Consequently, recent work has increasingly adopted LLM-as-a-Judge frameworks~\cite{zheng2023judging} and dimension-specific benchmarks for creative writing~\cite{zheng2025_cmlbench,kim2025_llm_creativity_eval}. However, LLM judges exhibit systematic biases, including same-family preference \cite{panickssery2024self,ye2024justice}, verbosity bias \cite{dubois2024length}, and limited agreement with expert evaluations \cite{fein2025litbench}. Existing mitigation strategies include cross-family judge panels \cite{verga2024poll}, structured deliberation \cite{chan2024chateval}, and evidence-based justifications \cite{jiang2025hamlet}; nevertheless, judge errors remain correlated across models \cite{kohli2026judgeseffectivevotescorrelated}, and conventional agreement measures can be distorted by skewed score distributions.

%% file: sections/3_problem.tex
\section{Problem Formulation}
\label{sec:problem-formulation}
We consider premise-conditioned plot generation with SLMs\footnote{We distinguish models by active parameter count rather than total parameters, and refer to models with at most 3B active parameters per token as SLMs, even when implemented as MoE architectures with larger total parameter counts.}: given a
concise, high-level premise specifying the narrative setting, genre, and thematic constraints (e.g., \texttt{a romantic comedy set in the modern tech startup era}), the model must produce a plot approaching professionally authored narrative quality. Drawing on computational narrative modeling and affective narratology \citep{chakrabarty2024art, hogan2011affective}, we assess quality along five Narrative Quality Dimensions (NQDs):
\emph{narrative coherence} (global logical consistency and causal connectivity), \emph{character development} (meaningful
character evolution), \emph{pacing} (distribution of narrative progression), \emph{tone consistency} (stylistic alignment), and
\emph{emotional turning points} (effectiveness of major affective transitions). Together, these dimensions span the structural, temporal, stylistic, character-centric, and affective aspects of narrative quality while keeping the evaluation space compact. Our objective is a structured
SLM-based workflow that generates premise-conditioned plots exhibiting strong performance across all NQDs.

%% file: sections/4_framework.tex
\section{Proposed Methodology}
\label{sec:proposed-model}
\begin{figure*}
    \centering
    \includegraphics[width=1\textwidth]{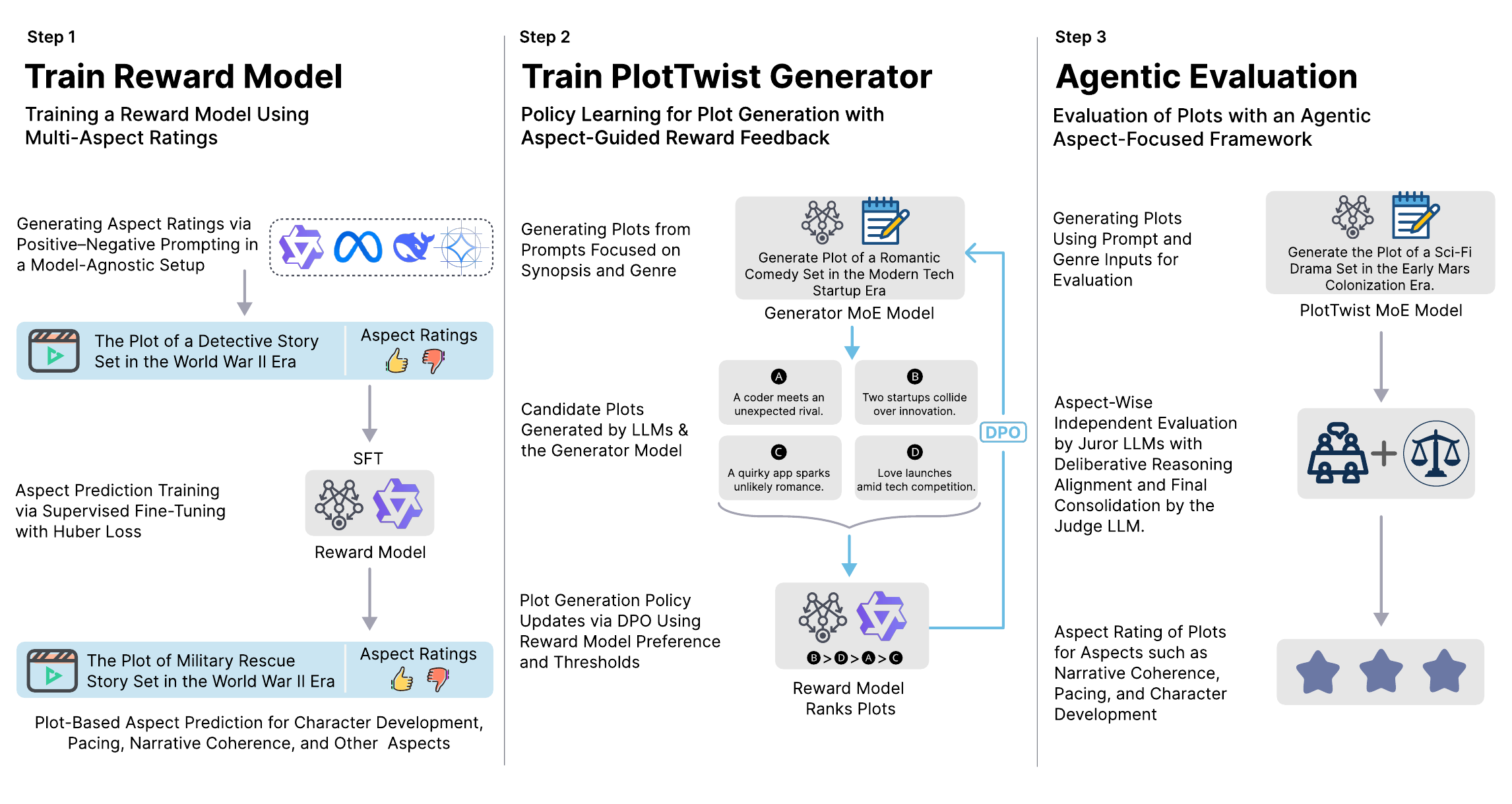}
    \caption{PlotTwist Framework}
    \label{fig:architecture}
\end{figure*}

As shown in Figure~\ref{fig:architecture}, \textsc{PlotTwist} comprises three modules: an Aspect Rating Reward Model that scores plots across the NQDs (Section~4.1), a preference-aligned Plot Generator (Section~4.2), and an independent Agentic Evaluation module with a cross-family jury (Section~4.3). 
\subsection{Aspect Rating Reward Model}
\label{sec:aspect-rating-estimation-model}
Our objective is to develop a reward model that produces aspect-level ratings and can be used to guide the plot generator model. We begin by constructing a dataset comprising plots paired with their corresponding ratings across the considered NQDs, and then fine-tune an LLM on this dataset to assign continuous-valued scores to plots. For a given plot-$p$ and aspect-$a$, we use $r_a(p)$ to denote the rating of plot-$p$ along aspect-$a$ in NQD.

\subsubsection{Aspect rating dataset construction}
As there are no existing datasets that provide fine-grained ratings of plots across the considered NQDs, we construct such a dataset synthetically using LLMs. To the best of our knowledge, the only widely available human-provided score is the IMDb rating for movies. However, this rating serves as a holistic assessment that encapsulates many creative elements simultaneously. Consequently, it cannot be used as a direct proxy for individual aspect-specific ratings, but it can serve as a coarse aggregate indicator of the overall quality of the plot. Note that IMDb ratings are not used in model training; they are employed solely for data curation and stratification purposes.
We begin by randomly sampling 5000 movies from the MovieLens~\cite{harper2015movielens} dataset spanning a broad range of IMDb ratings to ensure diversity. For each movie, we then scrape its corresponding plot from Wikipedia~\footnote{To meet the token output and computation requirements we select only movies with plots of at most 4000 words.}. Next, we employ LLMs to generate synthetic aspect-level ratings for each plot. We emphasize that LLMs are used solely in the reward model, while the final plot generator model, described in the following section, is implemented as an SLM.

For each plot $p$, we generate ratings for all aspects in NQDs using positive-negative prompting~\footnote{For reference, both positive and negative prompts for all aspects are given in the Appendix \ref{sec:appendix-reward-model-prompts}.} in a model agnostic setup, described below. This style of prompting mitigates the inherent positive bias often observed in LLMs~\cite{zheng2023judging} and yields stronger correlation with external indicators, enabling more accurate and balanced critique of plots.
We first take five LLMs, namely \texttt{Qwen-2.5-7B}, \texttt{Llama-3.3-70B}, \texttt{Llama-3.1-8B}, \texttt{DeepSeek-14B} and \texttt{Gemma-27B}, to avoid model bias in the aspect rating. We then prompt them with a plot to output a rating,  on a scale of 1-10, for each aspect by only considering the positives present in the plot along that aspect, denoted as $r_{a, m}^+(p)$ where $m$ identifies the LLM. Similarly, we prompt LLMs to output a rating, on a scale of 1-10, by only considering the negatives present in the plot along each aspect, denoted as $r_{a, m}^-(p).$ Note that, if an aspect is well captured in the plot then we expect $r_{a, m}^+(p)$ and $r_{a, m}^-(p)$ to be high and low respectively. 
Then, the final aspect rating of a plot is calculated as $r_a(p) = \sum_{m} \left( r_{a, m}^+(p) - r_{a, m}^-(p) \right)$. 
\begin{figure*}[!t]
    \centering
    \includegraphics[scale=0.4]{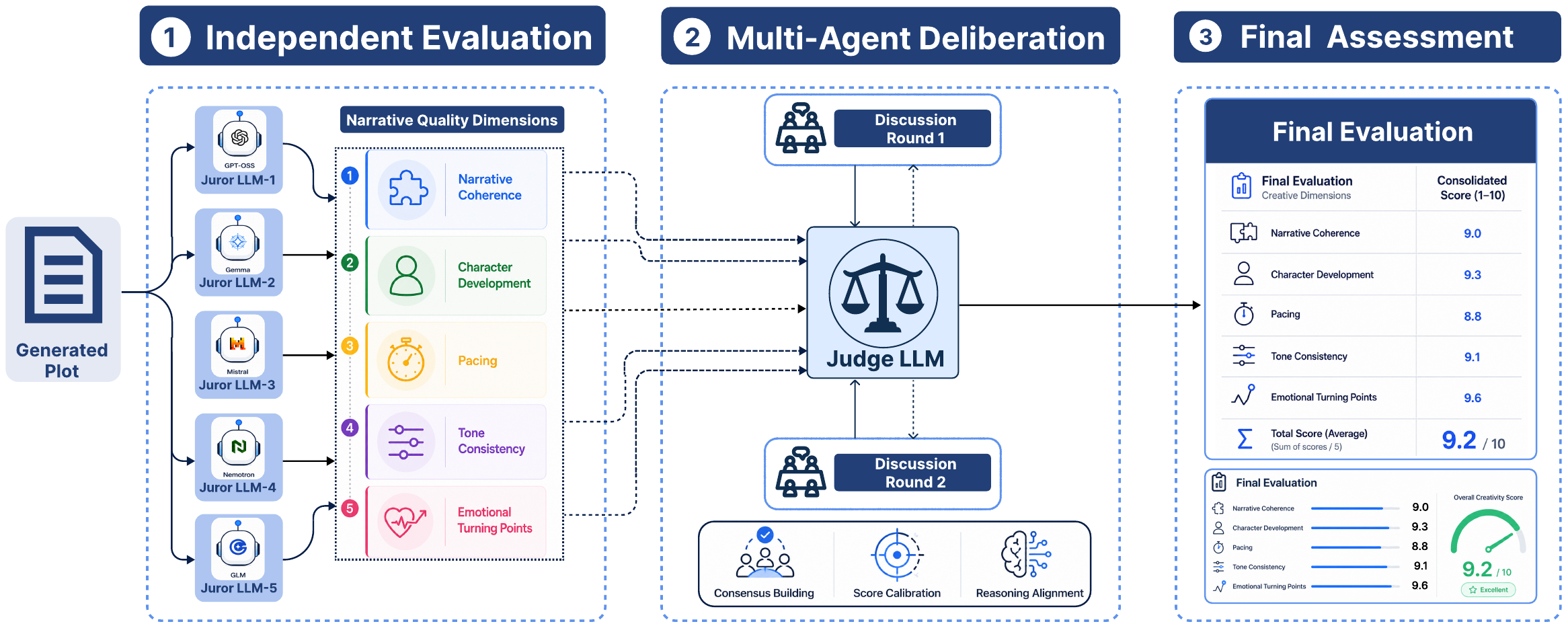}
    \caption{The Cross-Family Jury - Agentic Evaluation Framework}
    \label{fig:jury-framework}
\end{figure*}
\subsubsection{Supervised Fine-Tuning (SFT)}
We fine-tune Qwen-3-32B with 4-bit quantization on the aspect-rating dataset using a joint language-modeling and regression objective. The token-level Cross-Entropy (CE) loss supervises generation of the target response, whereas the Huber loss penalizes deviations between predicted and target aspect ratings. For an input sequence $x_{1:T}$, target sequence $y_{1:T}$, and model parameters $\theta$, the CE loss is defined as
\begin{equation*}
\small
\label{eq:CE-token-loss}
\mathcal{L}_{CE}(\theta) = -\frac{1}{T}\sum\nolimits_{t=1}^T \log p_\theta \left( y_t \mid x_{1:t} \right).
\end{equation*}
Let $r$ denote the residual between a predicted aspect rating and its target value. The Huber loss is defined as
\begin{equation*}
\small
\label{eq:Huber-loss}
\mathcal{L}_{\delta}(r) =
\begin{cases}
\frac{1}{2}r^2, & |r| \le \delta\\
\delta\left(|r|-\frac{1}{2}\delta\right), & |r|>\delta
\end{cases}
,
\end{equation*}
where we set $\delta=1$ in all experiments. 
\subsection{Plot Generator Model}
\label{sec:plot-generator-model}
We adopt Qwen-3-30B-A3B, an MoE architecture that enables increased model capacity and expert specialization while maintaining efficient inference \cite{fedus2022switch,shazeer2017outrageously}. 
Although the MoE model has a total parameter count of 30B, only 3B parameters are active per token, which classifies it as an SLM under our definition. To further align the model toward producing higher-quality plots, we employ Direct Preference Optimization (DPO), a Reinforcement Learning from Human/AI Feedback (RLHF/RLAIF) approach that directly optimizes preference objectives without requiring an explicit reward model or on-policy reinforcement learning. 
Given that the Qwen-3-30B-A3B model already exhibits strong instruction-following capabilities, we omit SFT for instruction alignment and focus exclusively on preference-based optimization.

For DPO training, we first construct a dataset of pairwise plot preferences (Figure~\ref{fig:dataset-curation-for-DPO} in Appendix~\ref{sec:appendix-dpo-curation}). Each sample in the dataset consists of a premise, a pair of plots generated under that premise, and a preference ordering between them. We generate premise descriptions for each of the 5,000 plots, as mentioned in Section~\ref{sec:aspect-rating-estimation-model}, using the Gemma-27B model. For each premise, we prompt the base Qwen-3-30B-A3B MoE model, along with several frontier models (Claude Sonnet 4, Gemini 2.0 Flash, and GPT-4.1), to generate plots conditioned on the same premise. All generated plots are evaluated using the reward model described in Section~\ref{sec:aspect-rating-estimation-model}, with aspect-level ratings averaged to obtain a final reward for each plot. To align our MoE model toward higher-quality plot generation, we retain only those samples in which a frontier model achieves the highest reward score, exceeds a score threshold of 8, and outperforms the next-best model by a margin of at least 0.5. Applying this procedure across all 5,000 premises yields 160 high-confidence preference samples. Although modest in size, this dataset is intentionally curated to ensure reliable preference signals, and prior work has shown that DPO can effectively leverage a small number of high-quality preference pairs. We subsequently perform DPO on the Qwen-3-30B-A3B model using this dataset, yielding the final plot generator model. Computational infrastructure details are provided in Appendix \ref{sec:appendix-compute}.

\subsection{Agentic Evaluation of Plots using Grounded Cross-Family Jury}
\label{agentic-evaluation}
Following plot generation, reliable validation of creative quality is essential. Although the aspect rating reward model provides structured supervision across the predefined NQDs, it remains a predictive model optimized for aspect-level signals derived from training data. Relying solely on this model risks evaluation bias, as it may reward patterns correlated with learned signals rather than reflect broader narrative soundness. Although direct evaluation by domain experts remains the gold standard for assessing plot quality, it is often impractical due to limited availability, time, and cost. Therefore, following prior work~\cite{teleki2025survey,kim2025_llm_creativity_eval}, we adopt an independent agentic evaluation framework, described below and illustrated in Figure~\ref{fig:jury-framework}.
Our agentic evaluation framework performs structured, multi-criteria assessment across the NQDs independently of the aspect rating reward model.  It employs a panel of five independent Juror LLMs to assess each generated plot across five NQDs. Each juror performs its evaluation independently to encourage diverse and unbiased assessments. Their individual judgments are then provided to a Judge LLM, which consolidates the evaluations through two rounds of structured deliberation with the jurors, resolving disagreements and calibrating scores where necessary. The Judge LLM produces the final consensus score for each NQD, yielding a robust, consistent, and transparent assessment while reducing the variance inherent in single-model evaluations.

To ensure evaluation that is reliable, consistent, and independent of the identity of the LLM applying it, each NQD is specified through a rubric of ten explicit, instruction-level criteria that translate abstract narrative concepts into concrete and observable failure modes; each criterion is scored on a $[0,1]$ scale, and the criterion scores are summed to yield an aspect score out of ten. Narrative coherence, for example, is assessed by identifying breakdowns in logical progression, causal relationships, and structural consistency, including plot
holes, contradictions, and incoherent world-building.
The remaining four dimensions are specified in the same style; the full rubrics and exact prompts appear in Appendix~\ref{sec:appendix-agentic-evaluation-prompts}.

%% file: sections/5_experiments.tex
\section{Experiments}
\label{sec:experiments}
We first validate both the proposed Aspect Rating Reward Model and the Cross-Family Jury-based Agentic Evaluation framework. Next, we evaluate the PlotTwist under varying conditions and introduce the baselines. Finally, we compare PlotTwist against these baselines and present ablation studies.
\subsection{Aspect Rating Reward Model and Agentic Evaluation: Validation}
\label{sec:validation-of-reward-and-agentic-evaluation}
We verify that both components assign higher scores to high-quality plots than to low-quality ones. Since IMDb ratings are an imperfect proxy for plot quality, we instead use critically acclaimed plots from the 101 Greatest Screenplays of All Time (GSAT) and critically panned plots from the Golden Raspberry Screenplay Awards (Razzies). Owing to the class imbalance (37 Razzie vs.\ 94 GSAT)\footnote{The imbalance arises because the Razzie Awards were established only in 1981, the Worst Screenplay award is not presented every year, and our 4,000-word plot-length filter further reduces the available Razzie pool.}, we employ repeated balanced subsampling, fixing the Razzie set and sampling an equal number of GSAT films over 1,000 runs. \\
\textbf{Aspect Rating Reward Model.} The reward model consistently separates GSAT and Razzie plots across all NQDs. GSAT plots achieve an overall mean score of $8.28$ compared with $7.21$ for Razzie plots, yielding a mean difference of $+1.07$ ($95\%$ CI $[0.96,1.19]$), with GSAT outperforming Razzie plots in $100\%$ of subsampling runs. The largest improvements are observed in pacing ($+1.41$), emotional turning points ($+1.13$), and narrative coherence ($+1.12$), while character development ($+0.75$) and tone consistency ($+0.96$) also exhibit clear separation. All NQDs show large effect sizes (Cohen's $d=2.77$--$4.68$), and Welch's $t$-tests confirm statistically significant differences across all dimensions, including the aggregate score ($t=9.69$, $p=3.41\times10^{-18}$). \\ 
\textbf{Agentic Evaluation.} The Cross-Family Jury-based agentic evaluator exhibits similar behavior, assigning consistently higher scores to GSAT plots across all NQDs. The overall mean score increases from $6.58$ (Razzie) to $8.33$ (GSAT), corresponding to a mean difference of $+1.75$ ($95\%$ CI $[1.66,1.85]$), with GSAT outperforming Razzie plots in $100\%$ of subsampling runs. The largest separations occur in narrative coherence ($+2.64$), tone consistency ($+1.99$), and pacing ($+1.91$), followed by character development ($+1.22$) and emotional turning points ($+0.98$). All NQDs exhibit very large effect sizes (Cohen's $d=4.36$--$9.34$), and Welch's $t$-tests confirm statistically significant differences across all dimensions, including the aggregate score ($t=16.54$, $p=2.96\times10^{-38}$). 

Together, these results demonstrate that both the proposed reward model and the Cross-Family Jury-based agentic evaluator reliably distinguish critically acclaimed from critically panned plots across all NQDs.

\subsection{Quality-Stratified Analysis of PlotTwist}
\label{sec:quality-stratified}
We consider 160 films partitioned into four IMDb-defined quality strata: Excellent (IMDb $>8$), Good ($7<$ IMDb $\leq8$), Mid ($6<$ IMDb $\leq7$), and Low (IMDb $\leq6$). For each source plot, we derive a premise using Gemma-27B and use it to condition PlotTwist. The final dataset contains 160 complete original--generated pairs (40 Excellent, 40 Good, 40 Mid, and 40 Low). Within each stratum, we compare paired original and generated plots across the five NQDs under the same Cross-Family Jury protocol as Section~\ref{sec:jury-evaluation}; uncertainty is estimated by resampling films rather than individual scores.

The overall improvement increases monotonically as source quality decreases: $+0.31$ points for Excellent films ($95\%$ CI $[0.09,0.57]$; paired Cohen's $d_z=0.38$), $+0.65$ for Good films ($[0.51,0.79]$; $d_z=1.40$), $+0.98$ for Mid films ($[0.81,1.17]$; $d_z=1.68$), and $+1.28$ for Low films ($[1.09,1.48]$; $d_z=2.01$). Within the Excellent stratum, the clearest gains occur in character development ($+0.50$, $[0.20,0.88]$) and narrative coherence ($+0.40$, $[0.13,0.72]$), whereas pacing remains unchanged within uncertainty ($-0.02$, $[-0.21,0.19]$). In the Low stratum, the largest improvements are observed in narrative coherence ($+1.82$) and tone consistency ($+1.40$), with generated plots outperforming their paired originals on at least $97.5\%$ of films across every NQD. This monotonic trend indicates quality-adaptive generation: PlotTwist primarily refines already strong narratives while progressively restructuring weaker ones (full per-stratum results are provided in Appendix~\ref{sec:appendix-quality-strata}).

\subsection{Baselines}
\label{sec:baselines}
We evaluate PlotTwist against baselines selected to provide a holistic comparison across three orthogonal axes: model scale, architectural design, and plot generation paradigm (see Appendix~\ref{sec:appendix-models-details} for a complete
overview of all models and their roles in the framework).
Closed-source frontier models (GPT-4.1\cite{openai2025gpt41}, Claude Sonnet~4\cite{anthropic2025sonnet4}, and Gemini 2.0 Flash\cite{gemini2024flash}) and large open-weight models (Llama-3.3-70B and Qwen-3-32B \cite{qwen2025qwen3}) test whether the structure of the framework can substitute for raw scale; instruction-tuned, reasoning-distilled, and reasoning-optimized models of comparable scale (Qwen-2.5-14B \cite{qwen2024qwen25}, DeepSeek-R1-14B \cite{deepseek2025r1}, Mistral Small 2501 \cite{mistral2025small3}, and Phi-4 Mini \cite{microsoft2025phi4mini}) separate the contribution of preference optimization from sparse activation and capacity; and two narrative-specific paradigms, Agents' Room \cite{huot2024agents} and WizardLM \cite{thebloke2023wizardlm}, situate PlotTwist among purpose-built story generation systems.

%% file: sections/5b_jury.tex
\subsection{PlotTwist Performance Evaluation}
\label{sec:jury-evaluation}
\input{tables/jury_means}
\textbf{Jury composition in Agentic Evaluation.} 
Since LLMs have been shown to exhibit bias toward outputs generated by models from the same family~\cite{panickssery2024self,ye2024justice}, we construct a jury comprising five open-weight LLMs from five model families different from our aspect rating reward model (Qwen): GPT-OSS-120B\cite{openai2025gptoss120bgptoss20bmodel}, Gemma-4-31B\cite{gemmateam2026gemma4technicalreport}, Mistral-Medium-3.5-128B, Nemotron-3-Super-120B\cite{nvidia2025nvidianemotron3efficient}, and GLM-4.6V \cite{5team2025glm45agenticreasoningcoding}, with Llama-3.3-70B serving as the Judge LLM. 
Each juror applies the same ten-criterion rubric described in Section~\ref{agentic-evaluation}, ensuring that only the underlying judge model varies across the panel (implementation details are provided in Appendix~\ref{sec:appendix-jury-details}). \\
\textbf{Evidence grounding.} Each criterion score must cite a span quoted verbatim from the plot being scored, turning written justifications \cite{jiang2025hamlet} into mechanically checkable claims: between $94\%$ and $99\%$ of the roughly $478{,}000$ quoted spans verify against the source text. Grounding is measured rather than enforced, avoiding the selection bias of dropping cells (Appendix~\ref{sec:appendix-jury-details}). 
\input{tables/jury_winrate}
We compare PlotTwist against the eleven baselines described in Section~\ref{sec:baselines} using the Cross-Family Jury Agentic Evaluation framework (Figure~\ref{fig:jury-framework}). Evaluation is performed on a held-out set of 160 premises sampled from the 5,000-premise dataset.
\\
\textbf{Mean scores.} PlotTwist and Claude Sonnet 4 achieve the highest overall jury mean scores, $8.36$ and $8.35$, respectively (Table~\ref{tab:jury_means}), with PlotTwist leading in narrative coherence, tone consistency, and pacing. Given the marginal difference in overall scores, we rely on the per-premise win rate (Table~\ref{tab:jury_winrate}) as the primary comparison metric. \\
\textbf{Win rates.} Following a pre-specified analysis plan, the primary endpoint is the per-premise paired win rate of PlotTwist against each baseline (ties counted as $0.5$), evaluated using premise-level bootstrap $95\%$ confidence intervals and Holm-corrected exact sign tests~\cite{holm1979simple} (Table~\ref{tab:jury_winrate}). PlotTwist achieves win rates of at least $0.95$ against every open-weight baseline, $0.94$ against Gemini 2.0 Flash, $0.91$ against GPT-4.1 ($95\%$ CI $[0.86, 0.95]$), and $0.73$ against Agents' Room, all remaining statistically significant after Holm correction for multiple comparisons ($p < 10^{-8}$). The GPT-4.1 result is consistent across jurors: each prefers PlotTwist on a majority of premises, and leave-one-juror-out analysis maintains a win rate of at least $0.85$ (Appendix~\ref{sec:appendix-jury-details}). In contrast, the comparison with Claude Sonnet 4 is inconclusive, with a win rate of $0.56$ ($95\%$ CI $[0.49, 0.64]$); neither the sign test ($p = 0.13$) nor the equivalence test~\cite{lakens2017equivalence} establishes superiority or equivalence. We therefore conclude only that PlotTwist is competitive with Claude Sonnet~4.\\
\textbf{Ablations.} The jury results show that PlotTwist's gains are not attributable to model scale, architecture, or generation paradigm. It outperforms similarly sized instruction-tuned models (Qwen2.5-14B: $7.62$; DeepSeek-R1 14B: $7.60$), the dense Qwen3-32B baseline ($7.90$) despite using roughly one-tenth as many active parameters per token (win rate $0.99$), and the multi-agent Agents' Room framework ($8.25$), achieving a higher overall mean ($8.36$) with a win rate of $0.73$ using a single model and inference pass. To isolate the effect of preference optimization, we evaluate the base Qwen-3-30B-A3B model before and after DPO using the development evaluator, observing an overall score increase from $8.03$ to $8.81$ (+$0.78$) after alignment on 160 high-confidence preference pairs. \\
\textbf{Reliability.} Raw inter-juror agreement is moderate (Krippendorff's $\alpha = 0.41$ \cite{krippendorff2004content}), largely reflecting calibration differences rather than ranking disagreement. Consistent with this, Gwet's AC2 is $0.89$ \cite{gwet2008computing}, recentering juror scores increases $\alpha$ to $0.64$, and system rankings are highly consistent across jurors (mean pairwise Spearman $0.97$, Kendall's $W = 0.98$). Accounting for correlated judgments, the panel provides an effective sample size of approximately $1.3$ independent judges rather than five~\cite{kohli2026judgeseffectivevotescorrelated}, motivating the per-judge and leave-one-judge-out analyses. \\
\textbf{Deliberation and final judge.} To probe shared error, score disagreements of at least two points between any pair of jurors are resolved through two round of structured deliberation, followed by adjudication from a held-out sixth-family judge, Llama-3.3-70B. Neither step alters the conclusions; accordingly, we report the independent first-round panel throughout. The system ranking is likewise robust to removing same-family judges, enforcing grounded evaluation, and all 113 analysis specifications (Appendix~\ref{sec:appendix-jury-details}).

%% file: tables/jury_means.tex
\begin{table*}[!ht]
\centering
\small
{\renewcommand{\arraystretch}{1.0}\setlength{\tabcolsep}{4.5pt}
\begin{tabular}{l|ccccc|c}
\toprule[1.1pt]
\rowcolor{gray!15}
\textbf{Model} & \shortstack{\textbf{Character} \\ \textbf{Development}} &
\shortstack{\textbf{Tone} \\ \textbf{Consistency}} & \textbf{Pacing} &
\shortstack{\textbf{Narrative} \\ \textbf{Coherence}} &
\shortstack{\textbf{Emotional} \\ \textbf{Turning Points}} & \textbf{Overall}\\
\midrule[1.1pt]
Qwen-2.5-14B & $7.41 \pm 0.59$ & $7.53 \pm 0.43$ & $7.52 \pm 0.31$ & $7.69 \pm 0.48$ & $7.95 \pm 0.39$ & $7.62 \pm 0.33$ \\
Claude Sonnet 4 & $\mathbf{8.30 \pm 0.41}$ & \underline{$8.23 \pm 0.35$} & \underline{$8.09 \pm 0.24$} & \underline{$8.49 \pm 0.38$} & $\mathbf{8.63 \pm 0.24}$ & \underline{$8.35 \pm 0.24$} \\
Gemini 2.0 Flash & $8.01 \pm 0.41$ & $7.98 \pm 0.32$ & $7.91 \pm 0.19$ & $8.27 \pm 0.32$ & $8.40 \pm 0.29$ & $8.11 \pm 0.24$ \\
GPT-4.1 & $7.88 \pm 0.52$ & $8.14 \pm 0.29$ & $7.99 \pm 0.19$ & $8.37 \pm 0.34$ & $8.41 \pm 0.28$ & $8.16 \pm 0.24$ \\
Llama-3.3-70B & $7.11 \pm 0.69$ & $7.39 \pm 0.55$ & $6.94 \pm 0.56$ & $6.89 \pm 0.75$ & $7.56 \pm 0.53$ & $7.18 \pm 0.48$ \\
DeepSeek-R1-14B & $7.29 \pm 0.66$ & $7.67 \pm 0.42$ & $7.53 \pm 0.31$ & $7.61 \pm 0.48$ & $7.89 \pm 0.41$ & $7.60 \pm 0.35$ \\
Phi-4 Mini & $6.58 \pm 0.94$ & $6.95 \pm 1.27$ & $6.27 \pm 1.55$ & $6.21 \pm 1.54$ & $6.88 \pm 1.07$ & $6.58 \pm 1.18$ \\
Qwen3-32B & $7.66 \pm 0.57$ & $8.10 \pm 0.34$ & $7.62 \pm 0.39$ & $7.83 \pm 0.57$ & $8.28 \pm 0.31$ & $7.90 \pm 0.33$ \\
Mistral Small 24B & $7.38 \pm 0.54$ & $7.70 \pm 0.42$ & $7.49 \pm 0.31$ & $7.57 \pm 0.53$ & $7.85 \pm 0.42$ & $7.60 \pm 0.34$ \\
Agents' Room & \underline{$8.15 \pm 0.38$} & $8.15 \pm 0.37$ & $7.97 \pm 0.37$ & $8.40 \pm 0.34$ & $8.57 \pm 0.38$ & $8.25 \pm 0.26$ \\
WizardLM-30B & $6.52 \pm 0.80$ & $7.07 \pm 0.55$ & $6.81 \pm 0.55$ & $6.53 \pm 0.70$ & $6.97 \pm 0.62$ & $6.78 \pm 0.51$ \\
\midrule
\rowcolor{gray!15}
\textbf{\textsc{PlotTwist}} & $8.11 \pm 0.60$ & $\mathbf{8.42 \pm 0.41}$ & $\mathbf{8.14 \pm 0.33}$ & $\mathbf{8.51 \pm 0.44}$ & \underline{$8.61 \pm 0.32$} & $\mathbf{8.36 \pm 0.35}$ \\
\bottomrule[1.1pt]
\end{tabular}
}
\caption{PlotTwist Performance Evaluation. Results are reported as mean $\pm$ standard deviation over 160 test premises using the Cross-Family Jury Agentic Evaluation framework. The best and second-best results in each column are shown in bold and underlined, respectively.}
\label{tab:jury_means}
\end{table*}

%% file: tables/jury_winrate.tex
\begin{table}[t]
\centering
\small
\setlength{\tabcolsep}{3.5pt}
\resizebox{\columnwidth}{!}{
\begin{tabular}{l|cc|cc}
\toprule[1.1pt]
\rowcolor{gray!15}
\textbf{Baseline} & \textbf{Mean} & \textbf{WR} & \textbf{$95\%$ CI} & \textbf{$p_{\text{Holm}}$} \\
\midrule[1.1pt]
Claude Sonnet 4   & 8.35 & $\mathbf{0.563}$ & [0.488, 0.638] & $0.13$ \\
Agents' Room      & 8.25 & $\mathbf{0.734}$ & [0.663, 0.800] & $4.4\times10^{-9}$ \\
GPT-4.1           & 8.16 & $\mathbf{0.906}$ & [0.863, 0.950] & $2.0\times10^{-27}$ \\
Gemini 2.0 Flash  & 8.11 & $\mathbf{0.938}$ & [0.900, 0.975] & $1.3\times10^{-32}$ \\
Qwen-3-32B        & 7.90 & $\mathbf{0.987}$ & [0.969, 1.000] & $2.1\times10^{-43}$ \\
Qwen-2.5-14B      & 7.62 & $\mathbf{0.950}$ & [0.913, 0.981] & $6.4\times10^{-35}$ \\
Mistral Small 24B & 7.60 & $\mathbf{0.994}$ & [0.981, 1.000] & $2.4\times10^{-45}$ \\
DeepSeek-R1-14B   & 7.60 & $\mathbf{0.994}$ & [0.981, 1.000] & $2.4\times10^{-45}$ \\
Llama-3.3-70B     & 7.18 & $\mathbf{0.988}$ & [0.969, 1.000] & $1.2\times10^{-43}$ \\
WizardLM-30B      & 6.78 & $\mathbf{0.994}$ & [0.981, 1.000] & $2.4\times10^{-45}$ \\
Phi-4 Mini        & 6.58 & $\mathbf{0.994}$ & [0.981, 1.000] & $2.4\times10^{-45}$ \\
\bottomrule[1.1pt]
\end{tabular}
}
\caption{Per-premise paired win rates of PlotTwist against each baseline under the Cross-Family Jury framework (ties counted as $0.5$), with premise-level bootstrap $95\%$ CIs and Holm-corrected exact sign tests. Mean: the baseline's overall mean from Table~\ref{tab:jury_means}.}
\label{tab:jury_winrate}
\end{table}

%% file: sections/6_conclusion.tex
\section{Conclusion}
\label{sec:conclusion}
We presented \textsc{PlotTwist}, a modular framework for premise-conditioned plot generation with SLMs. By combining aspect-rating reward modeling, preference optimization, and an independent cross-family jury agentic evaluation, \textsc{PlotTwist} enables efficient, high-quality plot generation while maintaining reliable and interpretable assessment of narrative quality. Extensive experiments demonstrate that PlotTwist consistently outperforms strong baselines despite using only 3B active parameters. Together, these findings underscore the value of structured preference-based alignment as a scalable and effective alternative to brute-force model scaling for creative text generation with limited-capacity language models.

%% file: sections/7_limitations.tex
\section*{\texorpdfstring{{Limitations}}{Limitations}}

As is standard in recent work on open-ended generation \cite{zheng2023judging}, our assessment of plot quality relies on LLM judges rather than expert annotation. The jury protocol is designed to address the known risks of this choice: judges are drawn from model families disjoint from the generator, every score must cite verifiable evidence, and reliability is reported alongside the results, with the reward model additionally validated against human-derived labels. LLM judges nevertheless agree only moderately with expert readers on creative writing \cite{chakrabarty2024art,fein2025litbench}, and two caveats are quantified in Section~\ref{sec:jury-evaluation}: the comparison with Claude Sonnet~4 remains statistically unresolved at $160$ premises, and correlated juror errors reduce the number of effectively independent opinions the panel provides. Our experiments are confined to English, synopsis-style plots scored on five craft-oriented dimensions, and the reward and preference data are constructed offline with the aid of larger models; extending the framework to other narrative forms, languages, and notions of quality such as novelty is a natural direction for future work.

%% file: sections/A_appendix.tex
\definecolor{cobaltlight}{HTML}{EDF1FA}
\definecolor{cobaltdark}{HTML}{4A6FA5}

\renewcommand{\arraystretch}{1.25}
\sloppy
\appendix
\section{Reward Model Prompts}
\label{sec:appendix-reward-model-prompts}

\begin{longtable}{p{0.96\linewidth}}

\toprule
\textbf{\large Aspect 1: Narrative Coherence} \\
\midrule

\textbf{\textcolor{cobaltdark}{Positive Prompt ($r^+_a$)}} \\

\rowcolor{cobaltlight}
You are a professional movie critic whose \textbf{only} output must be a \textbf{single JSON object} with exactly one integer field (0--10):\\
\texttt{Narrative Coherence:} \quad \textbf{Field Definition (Positive Focus):} \\

\rowcolor{cobaltlight}
Narrative clarity, logical plot progression, coherent world-building, strong cause--effect relationships, and well-integrated subplots. \\

\textbf{Strict output rules:} \\

\rowcolor{cobaltlight}
\begin{minipage}{\linewidth}
\begin{enumerate}
\item Output only a valid JSON object.
\item Include only \texttt{Narrative\_Coherence}.
\item Integer value from 0 to 10.
\item Score generously.
\end{enumerate}
\end{minipage} \\

\texttt{\#\#\# MoviePlot: \{ \}} \\
\midrule

\textbf{\textcolor{cobaltdark}{Negative Prompt ($r^-_a$)}} \\

\rowcolor{cobaltlight}
You are a professional movie critic whose \textbf{only} output must be a \textbf{single JSON object} with exactly one integer field (0--10):\\
\texttt{Narrative Coherence:} \quad \textbf{Field Definition (NegativeFocus):} \\

\rowcolor{cobaltlight}
Confusing storytelling, plot holes, inconsistent world-building, disconnected subplots, or illogical character decisions. \\

\textbf{Strict output rules:} \\

\rowcolor{cobaltlight}
\begin{minipage}{\linewidth}
\begin{enumerate}
\item Output only a valid JSON object.
\item Include only \texttt{Narrative\_Coherence}.
\item Integer value from 0 to 10.
\item 0 = no issues, 10 = severe issues.
\end{enumerate}
\end{minipage} \\

\texttt{\#\#\# MoviePlot: \{ \}} \\
\midrule

\textbf{\large Aspect 2: Emotional Turning Points} \\
\midrule

\textbf{\textcolor{cobaltdark}{Positive Prompt ($r^+_e$)}} \\

\rowcolor{cobaltlight}
You are a professional movie critic whose \textbf{only} output must be a \textbf{single JSON object} with exactly one integer field (0--10):\\

\texttt{Emotional Turning Points:} \quad \textbf{Field Definition (Positive Focus):} \\

\rowcolor{cobaltlight}
Powerful emotional moments, effective turning points, meaningful revelations, and emotionally satisfying narrative shifts. \\

\textbf{Strict output rules:} \\

\rowcolor{cobaltlight}
\begin{minipage}{\linewidth}
\begin{enumerate}
\item Output only JSON.
\item Include only \texttt{Emotions\_Turning\_Points}.
\item Integer 0--10.
\item Score generously.
\end{enumerate}
\end{minipage} \\

\texttt{\#\#\# MoviePlot: \{ \} \quad | \quad \#\#\# Review:} \\

\midrule

\textbf{\textcolor{cobaltdark}{Negative Prompt ($r^-_e$)}} \\

\rowcolor{cobaltlight}
\rowcolor{cobaltlight}
You are a professional movie critic whose \textbf{only} output must be a \textbf{single JSON object} with exactly one integer field (0--10):\\

\texttt{Emotional Turning Points:} \quad \textbf{Field Definition (Negative Focus):} \\

\rowcolor{cobaltlight}
Flat emotional arcs, forced turning points, unearned twists, or moments that fail to engage the audience. \\

\textbf{Strict output rules:} \\

\rowcolor{cobaltlight}
\begin{minipage}{\linewidth}
\begin{enumerate}
\item Output only JSON.
\item Include only \texttt{Emotions\_Turning\_Points}.
\item Integer 0--10.
\item 0 = no issues, 10 = severe issues.
\end{enumerate}
\end{minipage} \\

\texttt{\#\#\# MoviePlot: \{ \} \quad | \quad \#\#\# Review:} \\

\toprule
\textbf{\large Aspect 3: Tone Consistency} \\
\midrule

\textbf{\textcolor{cobaltdark}{Positive Prompt ($r^+_t$)}} \\

\rowcolor{cobaltlight}
You are a professional movie critic whose \textbf{only} output must be a \textbf{single JSON object} with exactly one integer field (0--10):\\
\texttt{Tone Consistency:} \quad \textbf{Field Definition (Positive Focus):} \\

\rowcolor{cobaltlight}
Successful maintenance of mood, atmosphere, and stylistic coherence throughout the story. Effective emotional consistency, well-maintained genre conventions, and smooth transitions between story beats. Intentional tonal shifts are rewarded when they serve the narrative purpose. \\

\textbf{Strict output rules:} \\

\rowcolor{cobaltlight}
\begin{minipage}{\linewidth}
\begin{enumerate}
\item Output only a valid JSON object.
\item Include only \texttt{Tone\_Consistency}.
\item Integer value from 0 to 10.
\item Score generously.
\end{enumerate}
\end{minipage} \\

\texttt{\#\#\# MoviePlot: \{ \} | \#\#\# Review:} \\
\midrule

\textbf{\textcolor{cobaltdark}{Negative Prompt ($r^-_t$)}} \\

\rowcolor{cobaltlight}
You are a professional movie critic whose \textbf{only} output must be a \textbf{single JSON object} with exactly one integer field (0--10):\\
\texttt{Tone Consistency:} \quad \textbf{Field Definition (Negative Focus):} \\

\rowcolor{cobaltlight}
Jarring mood shifts, inconsistent atmosphere, conflicting stylistic elements, genre incoherence, or awkward tonal transitions that disrupt immersion or emotional continuity. \\

\textbf{Strict output rules:} \\

\rowcolor{cobaltlight}
\begin{minipage}{\linewidth}
\begin{enumerate}
\item Output only a valid JSON object.
\item Include only \texttt{Tone\_Consistency}.
\item Integer value from 0 to 10.
\item 0 = no issues, 10 = severe issues.
\end{enumerate}
\end{minipage} \\

\texttt{\#\#\# MoviePlot: \{ \} | \#\#\# Review:} \\
\midrule
\textbf{\large Aspect 4: Character Development} \\
\midrule

\textbf{\textcolor{cobaltdark}{Positive Prompt ($r^+_c$)}} \\

\rowcolor{cobaltlight}
You are a professional movie critic whose \textbf{only} output must be a \textbf{single JSON object} with exactly one integer field (0--10):\\
\texttt{Character Development:} \quad \textbf{Field Definition (Positive Focus):} \\

\rowcolor{cobaltlight}
Compelling character arcs, meaningful growth, clear motivations, well-developed relationships, authentic character voices, and satisfying character journeys. Emphasis is placed on characters who evolve, learn, or change meaningfully over the course of the story. \\

\textbf{Strict output rules:} \\

\rowcolor{cobaltlight}
\begin{minipage}{\linewidth}
\begin{enumerate}
\item Output only a valid JSON object.
\item Include only \texttt{Character\_Development}.
\item Integer value from 0 to 10.
\item Score generously.
\end{enumerate}
\end{minipage} \\

\texttt{\#\#\# MoviePlot: \{ \} | \#\#\# Review:} \\
\midrule

\textbf{\textcolor{cobaltdark}{Negative Prompt ($r^-_c$)}} \\

\rowcolor{cobaltlight}
You are a professional movie critic whose \textbf{only} output must be a \textbf{single JSON object} with exactly one integer field (0--10):\\
\texttt{Character Development:} \quad \textbf{Field Definition (Negative Focus):} \\

\rowcolor{cobaltlight}
Weak or static character arcs, lack of growth, unclear motivations, poorly developed relationships, inconsistent character voices, or unsatisfying character journeys. Emphasis is placed on characters who remain static, act illogically, or fail to develop meaningfully. \\

\textbf{Strict output rules:} \\

\rowcolor{cobaltlight}
\begin{minipage}{\linewidth}
\begin{enumerate}
\item Output only a valid JSON object.
\item Include only \texttt{Character\_Development}.
\item Integer value from 0 to 10.
\item 0 = no issues, 10 = severe issues.
\end{enumerate}
\end{minipage} \\

\texttt{\#\#\# MoviePlot: \{ \} | \#\#\# Review:} \\
\midrule
\textbf{\large Aspect 5: Pacing} \\
\midrule

\textbf{\textcolor{cobaltdark}{Positive Prompt ($r^+_p$)}} \\

\rowcolor{cobaltlight}
You are a professional movie critic whose \textbf{only} output must be a \textbf{single JSON object} with exactly one integer field (0--10):\\
\texttt{Pacing:} \quad \textbf{Field Definition (Positive Focus):} \\

\rowcolor{cobaltlight}
Effective narrative rhythm, well-balanced scene progression, appropriate timing of plot events, and smooth transitions that maintain momentum and audience engagement. Emphasis is placed on pacing that supports tension, emotional beats, and story clarity. \\

\textbf{Strict output rules:} \\

\rowcolor{cobaltlight}
\begin{minipage}{\linewidth}
\begin{enumerate}
\item Output only a valid JSON object.
\item Include only \texttt{Pacing}.
\item Integer value from 0 to 10.
\item Score generously.
\end{enumerate}
\end{minipage} \\

\texttt{\#\#\# MoviePlot: \{ \} | \#\#\# Review:} \\
\midrule

\textbf{\textcolor{cobaltdark}{Negative Prompt ($r^-_p$)}} \\

\rowcolor{cobaltlight}
You are a professional movie critic whose \textbf{only} output must be a \textbf{single JSON object} with exactly one integer field (0--10):\\
\texttt{Pacing:} \quad \textbf{Field Definition (Negative Focus):} \\

\rowcolor{cobaltlight}
Uneven or inconsistent pacing, excessive slowdowns or rushed segments, poorly timed plot events, unnecessary filler scenes, or abrupt transitions that disrupt narrative flow or emotional impact. \\

\textbf{Strict output rules:} \\

\rowcolor{cobaltlight}
\begin{minipage}{\linewidth}
\begin{enumerate}
\item Output only a valid JSON object.
\item Include only \texttt{Pacing}.
\item Integer value from 0 to 10.
\item 0 = no issues, 10 = severe issues.
\end{enumerate}
\end{minipage} \\

\texttt{\#\#\# MoviePlot: \{ \} | \#\#\# Review:} \\
\midrule

\end{longtable}

\newpage

\section{\texorpdfstring{DPO Data Curation}{DPO Data Curation}}
\label{sec:appendix-dpo-curation}
\begin{figure}[ht]
    \centering
    \includegraphics[width=\textwidth]{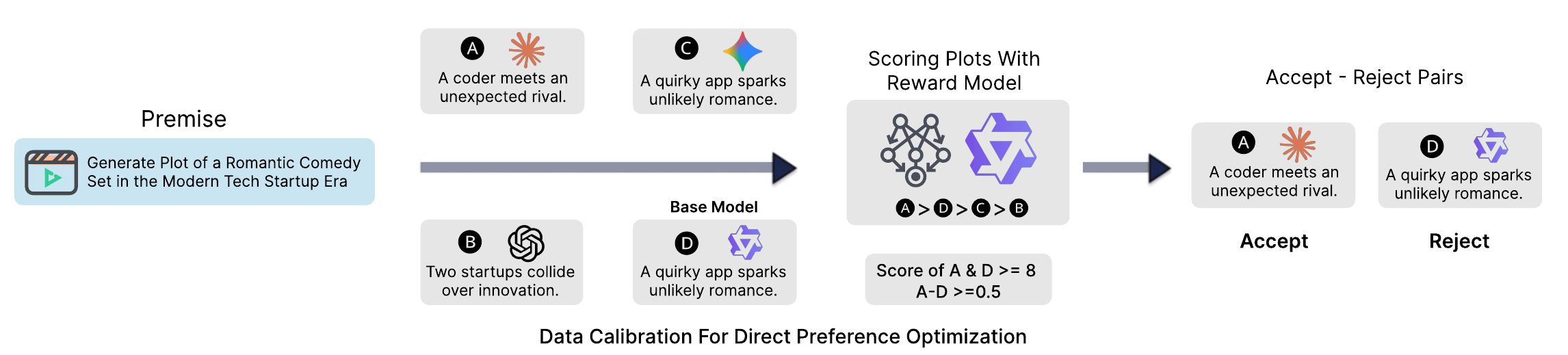}
    \caption{Dataset curation for performing DPO}
    \label{fig:dataset-curation-for-DPO}
\end{figure}

Figure~\ref{fig:dataset-curation-for-DPO} illustrates the preference-pair curation procedure described in Section~\ref{sec:plot-generator-model}.

\newpage

\section{Agentic Evaluation System Prompts}
\label{sec:appendix-agentic-evaluation-prompts}
\sloppy
\begin{longtable}{p{0.96\linewidth}}

\toprule
\textbf{\large System Prompt 1: Narrative Coherence Evaluation} \\
\midrule

\textbf{Task Overview} \\

\rowcolor{cobaltlight}
Evaluate a movie plot's narrative structure and logical consistency using a 10-criteria framework. Assign precise numerical scores reflecting coherence quality. \\

\textbf{Evaluation Methodology} \\

\rowcolor{cobaltlight}
Each criterion is scored from 0--1 (increments of 0.1 allowed). Scores are summed for a total out of 10. \\

\textbf{Scoring Criteria} \\

\rowcolor{cobaltlight}
\textbf{Plot Structure and Logic (4 points)} \\

\rowcolor{cobaltlight}
\begin{tabular}{@{}p{0.47\linewidth} p{0.47\linewidth}@{}}
\textbf{1. Plot Progression} &
\textbf{2. Causal Connectivity} \\[-2pt]
Logical beginning--middle--end flow. &
Events arise naturally from prior actions. \\[6pt]

\textbf{3. Plot Integrity} &
\textbf{4. Conflict Focus} \\[-2pt]
No plot holes or contradictions. &
A sustained central conflict drives the story. \\
\end{tabular} \\

\rowcolor{cobaltlight}
\textbf{Character Integration (3 points)} \\
5. Protagonist Consistency \quad
6. Supporting Character Function \quad
7. Resolution Authenticity \\

\rowcolor{cobaltlight}
\textbf{Narrative Flow and Unity (3 points)} \\
8. Pacing Appropriateness \quad
9. Thematic Integration \quad
10. Tonal Consistency \\

\textbf{Output Format} \\

\rowcolor{cobaltlight}
\begin{tabular}{@{}p{0.47\linewidth} p{0.47\linewidth}@{}}
\texttt{1. Plot Progression: X.X} &
\texttt{2. Causal Connectivity: X.X} \\
\texttt{3. Plot Integrity: X.X} &
\texttt{4. Conflict Focus: X.X} \\
\texttt{5. Protagonist Consistency: X.X} &
\texttt{6. Supporting Character Function: X.X} \\
\texttt{7. Resolution Authenticity: X.X} &
\texttt{8. Pacing Appropriateness: X.X} \\
\texttt{9. Thematic Integration: X.X} &
\texttt{10. Tonal Consistency: X.X} \\
\end{tabular} \\[-2pt]

\rowcolor{cobaltlight}
\texttt{TOTAL: X.X/10} \\

\midrule

\newpage
\textbf{\large System Prompt 2: Emotional Turning Point Evaluation} \\
\midrule

\textbf{Task Overview} \\

\rowcolor{cobaltlight}
Identify and evaluate the primary emotional turning point of the narrative using a 10-criteria framework focused on emotional impact and character change. \\

\textbf{Scoring Criteria} \\

\rowcolor{cobaltlight}
\textbf{Conflict \& Character Foundation (4 points)} \\

\rowcolor{cobaltlight}
\begin{tabular}{@{}p{0.47\linewidth} p{0.47\linewidth}@{}}
\textbf{1. Conflict Resolution} &
\textbf{2. Character Believability} \\[-2pt]
Addresses or reframes central conflict. &
Emotion aligns with established arc.\\[-2pt]

\textbf{3. Character Transformation} &
\textbf{4. Emotional Satisfaction} \\[-2pt]
Meaningful internal change. &
Emotionally resonant payoff. \\
\end{tabular} \\

\rowcolor{cobaltlight}
\textbf{Narrative Construction (3 points)} \\
5. Narrative Causality \quad
6. Thematic Crystallization \quad
7. Relationship Impact \\

\rowcolor{cobaltlight}
\textbf{Technical \& Structural Elements (3 points)} \\
8. Cinematic Execution \quad
9. Structural Necessity \quad
10. Audience Alignment \\

\textbf{Output Format} \\

\rowcolor{cobaltlight}
\begin{tabular}{@{}p{0.47\linewidth} p{0.47\linewidth}@{}}
\texttt{1. Conflict Resolution: X.X} &
\texttt{2. Character Believability: X.X} \\
\texttt{3. Character Transformation: X.X} &
\texttt{4. Emotional Satisfaction: X.X} \\
\texttt{5. Narrative Causality: X.X} &
\texttt{6. Thematic Crystallization: X.X} \\
\texttt{7. Relationship Impact: X.X} &
\texttt{8. Cinematic Execution: X.X} \\
\texttt{9. Structural Necessity: X.X} &
\texttt{10. Audience Alignment: X.X} \\
\end{tabular} \\[-2pt]

\rowcolor{cobaltlight}
\texttt{TOTAL: X.X/10} \\

\midrule
\end{longtable}

\begin{longtable}{p{0.96\linewidth}}

\toprule
\textbf{\large System Prompt 3: Character Development Evaluation} \\
\midrule

\textbf{Task Overview} \\

\rowcolor{cobaltlight}
Evaluate protagonist character development using a 10-criteria framework assessing motivation, arc progression, and narrative function. \\

\textbf{Scoring Criteria} \\

\rowcolor{cobaltlight}
\textbf{Core Character Elements (4 points)} \\

\rowcolor{cobaltlight}
\begin{tabular}{@{}p{0.47\linewidth} p{0.47\linewidth}@{}}
\textbf{1. Motivation Clarity} &
\textbf{2. Behavioral Consistency} \\[-2pt]
Clear goals and desires. &
Actions align with personality. \\[6pt]

\textbf{3. Character Arc} &
\textbf{4. Psychological Depth} \\[-2pt]
Believable transformation. &
Emotional and psychological complexity. \\
\end{tabular} \\

\rowcolor{cobaltlight}
\textbf{Character Foundation (3 points)} \\
5. Backstory Integration \quad
6. Audience Connection \quad
7. Character Distinctiveness \\

\rowcolor{cobaltlight}
\textbf{Narrative Function (3 points)} \\
8. Relationship Dynamics \quad
9. Plot Agency \quad
10. Thematic Alignment \\

\textbf{Output Format} \\

\rowcolor{cobaltlight}
\begin{tabular}{@{}p{0.47\linewidth} p{0.47\linewidth}@{}}
\texttt{1. Motivation Clarity: X.X} &
\texttt{2. Behavioral Consistency: X.X} \\
\texttt{3. Character Arc: X.X} &
\texttt{4. Psychological Depth: X.X} \\
\texttt{5. Backstory Integration: X.X} &
\texttt{6. Audience Connection: X.X} \\
\texttt{7. Character Distinctiveness: X.X} &
\texttt{8. Relationship Dynamics: X.X} \\
\texttt{9. Plot Agency: X.X} &
\texttt{10. Thematic Alignment: X.X} \\
\end{tabular} \\[-2pt]

\rowcolor{cobaltlight}
\texttt{TOTAL: X.X/10} \\

\midrule

\newpage
\textbf{\large System Prompt 4: Pacing Analysis Evaluation} \\
\midrule

\textbf{Task Overview} \\

\rowcolor{cobaltlight}
Assess narrative pacing using a 10-criteria framework measuring rhythm, momentum, and emotional timing. \\

\textbf{Scoring Criteria} \\

\rowcolor{cobaltlight}
\begin{tabular}{@{}p{0.47\linewidth} p{0.47\linewidth}@{}}
\textbf{1. Premise Establishment Speed} &
\textbf{2. Structural Foundation} \\

\textbf{3. Pacing Consistency} &
\textbf{4. Event Frequency} \\

\textbf{5. Scene Purposefulness} &
\textbf{6. Tension Management} \\

\textbf{7. Transition Quality} &
\textbf{8. Emotional Beat Timing} \\

\textbf{9. Climax Timing} &
\textbf{10. Genre--Tone Alignment} \\
\end{tabular} \\

\textbf{Output Format} \\

\rowcolor{cobaltlight}
\begin{tabular}{@{}p{0.47\linewidth} p{0.47\linewidth}@{}}
\texttt{1. Premise Establishment Speed: X.X} &
\texttt{2. Structural Foundation: X.X} \\
\texttt{3. Pacing Consistency: X.X} &
\texttt{4. Event Frequency: X.X} \\
\texttt{5. Scene Purposefulness: X.X} &
\texttt{6. Tension Management: X.X} \\
\texttt{7. Transition Quality: X.X} &
\texttt{8. Emotional Beat Timing: X.X} \\
\texttt{9. Climax Timing: X.X} &
\texttt{10. Genre--Tone Alignment: X.X} \\
\end{tabular} \\[-2pt]

\rowcolor{cobaltlight}
\texttt{TOTAL: X.X/10} \\

\midrule
\end{longtable}
\begin{longtable}{p{0.96\linewidth}}

\toprule
\textbf{\large System Prompt 5: Tone Consistency Evaluation} \\
\midrule

\textbf{Task Overview} \\

\rowcolor{cobaltlight}
Evaluate tonal coherence using a 10-criteria framework assessing atmosphere, stylistic unity, and emotional continuity. \\

\textbf{Scoring Criteria} \\

\rowcolor{cobaltlight}
\begin{tabular}{@{}p{0.47\linewidth} p{0.47\linewidth}@{}}
\textbf{1. Initial Atmosphere Establishment} &
\textbf{2. Scene-to-Scene Consistency} \\

\textbf{3. Tonal Relief Integration} &
\textbf{4. Earned Tone Shifts} \\

\textbf{5. Dialogue Style Consistency} &
\textbf{6. Visual Reinforcement} \\

\textbf{7. Stakes Alignment} &
\textbf{8. Comedy/Drama Balance} \\

\textbf{9. Ending Consistency} &
\textbf{10. Motif and Symbol Unity} \\
\end{tabular} \\

\textbf{Output Format} \\

\rowcolor{cobaltlight}
\begin{tabular}{@{}p{0.47\linewidth} p{0.47\linewidth}@{}}
\texttt{1. Initial Atmosphere Establishment: X.X} &
\texttt{2. Scene-to-Scene Consistency: X.X} \\
\texttt{3. Tonal Relief Integration: X.X} &
\texttt{4. Earned Tone Shifts: X.X} \\
\texttt{5. Dialogue Style Consistency: X.X} &
\texttt{6. Visual Reinforcement: X.X} \\
\texttt{7. Stakes Alignment: X.X} &
\texttt{8. Comedy/Drama Balance: X.X} \\
\texttt{9. Ending Consistency: X.X} &
\texttt{10. Motif and Symbol Unity: X.X} \\
\end{tabular} \\[-2pt]

\rowcolor{cobaltlight}
\texttt{TOTAL: X.X/10} \\

\bottomrule
\end{longtable}

\lstdefinestyle{promptjson}{%
  basicstyle=\ttfamily\footnotesize,
  breaklines=true,
  columns=fullflexible,
  keepspaces=true,
  showstringspaces=false,
  xleftmargin=0.5em,
  frame=none,
  aboveskip=4pt,
  belowskip=8pt,
}

\begin{longtable}{p{0.96\linewidth}}
\toprule
\textbf{\large System Prompt 6: Grounded Output Contract (Independent Scoring)} \\
\midrule

\textbf{Applied in} \\

\rowcolor{cobaltlight}
Round independent scoring. Appended verbatim to each of the five rubrics above, so every juror returns one structured, evidence-grounded record per NQD. \\

\textbf{Instruction (verbatim)} \\

\rowcolor{cobaltlight}
Return ONLY a single JSON object and nothing else, with the exact shape below. List every rubric criterion in order. Every \texttt{evidence\_quote} must be copied verbatim from the plot text, not paraphrased. Do not add any text outside the JSON object. \\

\textbf{Output schema} \\
\bottomrule
\end{longtable}

\begin{lstlisting}[style=promptjson]
{"nqd": "<the nqd key>",
 "criteria": [{"name": "<criterion name from the rubric above>",
               "score_0_1": <number between 0 and 1>,
               "evidence_quote": "<a span copied verbatim from the plot>",
               "rationale": "<one sentence>"}],
 "total_0_10": <sum of the criterion scores, between 0 and 10>,
 "overall_rationale": "<two sentences>"}
\end{lstlisting}

\begin{longtable}{p{0.96\linewidth}}
\toprule
\textbf{\large System Prompt 7: Deliberation Instruction (Juror Revision)} \\
\midrule

\textbf{Applied in} \\

\rowcolor{cobaltlight}
Both deliberation rounds, on cells the independent panel contests (a spread of at least two points between juror scores). Appended to the same rubric; each juror sees its own prior assessment and the anonymized, order-shuffled arguments and cited spans of the other jurors, with peer scores hidden to prevent vote-matching. \\

\textbf{Instruction (verbatim)} \\

\rowcolor{cobaltlight}
You already assessed this plot on the dimension defined by the rubric above. Below the plot you are shown (a) your own prior assessment and (b) the anonymized arguments of the other independent jurors, with the verbatim spans they cited. Reconsider your assessment in light of their EVIDENCE. Change your score only if a peer cites stronger grounded evidence than you did; do NOT change it to match a majority or because a peer sounds confident. If your original judgement still holds, keep your score and say why. Stay blind to anything outside the plot text (there is no premise, genre label, or reference answer). \\

\textbf{Output schema} \\
\bottomrule
\end{longtable}

\begin{lstlisting}[style=promptjson]
{"nqd": "<the nqd key>",
 "revised_total_0_10": <0 to 10>,
 "changed": <true|false>,
 "criteria": [{"name": "<criterion>",
               "score_0_1": <0 to 1>,
               "evidence_quote": "<verbatim span from the plot>",
               "rationale": "<one sentence>"}],
 "rebuttal": "<2 to 3 sentences: what you kept or moved, and which grounded peer argument did or did not change your mind>"}
\end{lstlisting}

\begin{longtable}{p{0.96\linewidth}}
\toprule
\textbf{\large System Prompt 8: Consolidation Instruction (Judge LLM)} \\
\midrule

\textbf{Applied in} \\

\rowcolor{cobaltlight}
Final consolidation by the held-out Judge LLM (Llama-3.3-70B) after the two deliberation rounds, once per contested NQD. The judge reads the plot, the anonymized independent assessments, and both rounds of revisions before issuing one consolidated score. \\

\textbf{Instruction (verbatim)} \\

\rowcolor{cobaltlight}
You are the presiding judge. Below the plot you are shown the anonymized, grounded assessments of several independent jurors on the dimension defined by the rubric above, each with the verbatim spans it cited. Produce ONE final score for this dimension by WEIGHING the arguments on how well each is grounded in evidence actually present in the plot, not by averaging the jurors and not by counting how many agree. Discount any claim whose cited span does not support it or is absent from the plot. Name the main point of disagreement and how the evidence resolves it. Stay blind to anything outside the plot text. \\

\textbf{Output schema} \\
\bottomrule
\end{longtable}

\begin{lstlisting}[style=promptjson]
{"nqd": "<the nqd key>",
 "final_total_0_10": <0 to 10>,
 "key_disagreement": "<the main split and how you resolved it>",
 "synthesis_rationale": "<3 to 4 sentences citing the decisive grounded evidence>",
 "criteria": [{"name": "<criterion>",
               "score_0_1": <0 to 1>,
               "evidence_quote": "<verbatim span>",
               "rationale": "<one sentence>"}]}
\end{lstlisting}

\section{Plot Generation with PlotTwist}
\label{sec:appendix-plot-generation}
\begin{longtable}{p{0.96\linewidth}}
\toprule
\textbf{\large Plot Generation with PlotTwist} \\
\midrule

\textbf{Task Description} \\

\rowcolor{cobaltlight}
The following example demonstrates long-form creative movie plot generation under the \textit{PlotTwist} setting. The task emphasizes narrative coherence, concealed truths, emotional stakes, and morally consequential decision-making, serving as a qualitative illustration of model behavior. \\

\textbf{Premise} \\

\rowcolor{cobaltlight}
Generate a movie plot that follows a dedicated detective as he investigates a murder that implicates his beloved, forcing him to navigate a web of concealed truths, blackmail, and the desperate lengths one will go to protect a dark secret. \\

\textbf{PlotTwist Generated Plot} \\

\rowcolor{cobaltlight}
Detective Aris Thorne, a man whose sharp eyes and even sharper instincts have solved countless cases but whose heart remains scarred by the unsolved murder of his wife Elara five years prior, finds his world shattered when the body of City Councilman Silas Croft---a powerful figure known for his ruthless political maneuvering and rumored ties to the city's underbelly---is discovered in his opulent penthouse with a single, precise stab wound to the heart.

The evidence points damningly toward Lena Voss, Aris's former lover and now a devoted social worker fighting for the city's forgotten children. Witnesses report a heated argument between Lena and Croft hours before his death, and a fragmented security recording captures her leaving the building, her face pale with fear. Aris, torn between duty and lingering love, cannot believe she is capable of murder---yet the physical evidence is overwhelming. A rare antique switchblade, identical to one Aris gifted Lena years earlier, is found clutched in Croft's hand. Even more troubling is a cryptic encrypted file on Croft's laptop, labeled \texttt{Project Chimera}, implicating Lena in a scheme involving missing children.

As Aris investigates further, the case reveals a labyrinth of concealed truths. Croft was not merely corrupt; he was the linchpin of an extensive blackmail network exploiting vulnerable families. The encrypted files reveal that Croft had been threatening Lena with exposure---not only of her past, but of evidence tied directly to Elara's death. Aris had long believed Elara's death was a tragic accident and had quietly suppressed details to protect her reputation.

The blackmail was never solely about Lena---it was about Aris himself. Lena ultimately murdered Croft in desperation to stop him from exposing the truth, protect the children he was trafficking, and prevent Aris's buried past from destroying them both. Aris finds himself ensnared between the city's elite, an ambitious District Attorney eager to frame Lena, Croft's remaining associates intent on silencing her, and his own conscience, which recognizes that Lena's act was born of sacrifice rather than malice.

Aris uncovers the final truth: Elara had been investigating Croft's operations as a journalist and was killed in a staged car accident when she came too close. In a climactic confrontation at the rain-lashed docks, Aris dismantles the remnants of Croft's network, but the true reckoning is internal. Choosing love over career, he confesses to the murder to protect Lena, sacrificing his badge and freedom.

As rain washes over the city, Aris walks away from everything he once was, the weight of concealed truths finally lifted. What remains is the memory of Lena---not as a suspect, but as the woman he loved---and the understanding that truth and love, when intertwined, can demand a devastating cost. \\

\bottomrule
\end{longtable}

\newpage
\section{Models and Their Roles in the PlotTwist Framework}
\label{sec:appendix-models-details}
\noindent Overview of all language models employed across training, evaluation, and baseline comparison stages of the PlotTwist system.

\vspace{10pt}

\definecolor{cobaltdark}{RGB}{0, 71, 171}
\definecolor{cobaltlight}{RGB}{230, 240, 255}

\begin{longtable}{p{\linewidth}}

\toprule
\textbf{\large PlotTwist Core Components} \\
\midrule
\textbf{\textcolor{cobaltdark}{Qwen-3-30B-A3B (MoE) --- PlotTwist Plot Generator}} \\
\rowcolor{cobaltlight}
MoE backbone with \textbf{3B active parameters}; preference-aligned via \textbf{Direct Preference Optimization (DPO)} to generate high-quality premise-conditioned plots. Only 3B parameters are active per token despite a total count of 30B, classifying it as an SLM under the paper's definition. \\
\midrule
\textbf{\textcolor{cobaltdark}{Qwen-3-32B (4-bit) --- PlotTwist Aspect Rating Reward Model}} \\
\rowcolor{cobaltlight}
Fine-tuned via \textbf{regression-aware SFT} using a weighted combination of \textbf{cross-entropy loss} and \textbf{Huber loss} to predict continuous aspect-level narrative quality scores across all five NQDs. \\
\midrule
\textbf{\textcolor{cobaltdark}{Cross-Family Jury --- PlotTwist Agentic Evaluation Module}} \\
\rowcolor{cobaltlight}
Independent post-hoc evaluation module operating separately from the training pipeline. A panel of five open-weight \textbf{Juror LLMs} drawn from five model families disjoint from the Qwen family of the generator and reward model: \textbf{GPT-OSS-120B}, \textbf{Gemma-4-31B}, \textbf{Mistral-Medium-3.5-128B}, \textbf{Nemotron-3-Super-120B-A12B}, and \textbf{GLM-4.6V}. Each juror independently scores all five \textbf{NQDs} using ten-criterion, weakness-focused rubrics with verbatim evidence grounding. Score disagreements of at least two points are resolved through \textbf{two rounds of structured deliberation}, consolidated by a held-out \textbf{Judge LLM}, \textbf{Llama-3.3-70B}, drawn from a sixth model family disjoint from both the panel and the generator. \\
\bottomrule

\toprule
\textbf{\large Ensemble Models for Positive-Negative Aspect Rating} \\
\midrule
\textbf{\textcolor{cobaltdark}{Qwen-2.5-7B \quad Llama-3.3-70B \quad Llama-3.1-8B \quad DeepSeek-14B \quad Gemma-27B}} \\
\rowcolor{cobaltlight}
Five-model ensemble used to generate \textbf{synthetic aspect-level ratings} via positive--negative prompting. Each model outputs both a positive score $r^+_{a,m}(p)$ and a negative score $r^-_{a,m}(p)$ per aspect, aggregated as:
\[
r_a(p) = \sum_m \left( r^+_{a,m}(p) - r^-_{a,m}(p) \right)
\]
Model diversity across the ensemble mitigates individual model bias in the rating construction process. \textbf{Gemma-27B} additionally generates premise descriptions from movie plots for DPO dataset curation. \\
\bottomrule

\toprule
\textbf{\large DPO Candidate Plot Generators} \\
\midrule
\textbf{\textcolor{cobaltdark}{GPT-4.1}} \\
\rowcolor{cobaltlight}
Frontier model used to generate candidate plots per premise for reward-model scoring and \textbf{DPO preference pair construction}. Retained as accepted plot only when it achieves the highest reward score ($\geq 8$) and outperforms the next-best model by a margin of at least 0.5. \\
\midrule
\textbf{\textcolor{cobaltdark}{Claude Sonnet 4}} \\
\rowcolor{cobaltlight}
Frontier model used to generate candidate plots per premise for reward-model scoring and \textbf{DPO preference pair construction}. Retained as accepted plot only when it achieves the highest reward score ($\geq 8$) and outperforms the next-best model by a margin of at least 0.5. \\
\midrule
\textbf{\textcolor{cobaltdark}{Gemini 2.0 Flash}} \\
\rowcolor{cobaltlight}
Frontier model used to generate candidate plots per premise for reward-model scoring and \textbf{DPO preference pair construction}. Retained as accepted plot only when it achieves the highest reward score ($\geq 8$) and outperforms the next-best model by a margin of at least 0.5. \\
\bottomrule

\toprule
\textbf{\large Baselines: Model Scale} \\
\midrule
\textbf{\textcolor{cobaltdark}{Llama-3.3-70B}} \\
\rowcolor{cobaltlight}
Large open-weight baseline. Tests performance at high parameter count \textbf{without task-specific alignment}, providing an upper-bound reference for scale alone. \\
\midrule
\textbf{\textcolor{cobaltdark}{Qwen3-32B (Dense)}} \\
\rowcolor{cobaltlight}
Dense baseline from the same model family as the PlotTwist generator. Used to \textbf{isolate performance gains attributable to DPO alignment} from those due to the MoE architecture. \\
\midrule
\textbf{\textcolor{cobaltdark}{Qwen2.5-7B-Instruct \quad Qwen2.5-14B-Instruct}} \\
\rowcolor{cobaltlight}
Small instruction-tuned baselines used for \textbf{scale ablation}, confirming that PlotTwist's gains arise from methodology rather than model size. \\
\bottomrule

\toprule
\textbf{\large Baselines: Architectural Design} \\
\midrule
\textbf{\textcolor{cobaltdark}{DeepSeek-R1 14B}} \\
\rowcolor{cobaltlight}
MoE and reasoning-oriented baseline. Contrasts \textbf{sparse activation without preference alignment}, isolating the contribution of DPO from that of expert routing. \\
\midrule
\textbf{\textcolor{cobaltdark}{Phi-4 Mini Instruct}} \\
\rowcolor{cobaltlight}
Compact reasoning-optimized model baseline. Evaluates whether \textbf{small reasoning-focused architectures} can match structured preference-aligned generation. \\
\midrule
\textbf{\textcolor{cobaltdark}{Mistral Small 2501 24B}} \\
\rowcolor{cobaltlight}
Reasoning-optimized baseline assessing \textbf{structured temporal progression} and narrative coherence in mid-scale models. \\
\bottomrule

\toprule
\textbf{\large Baselines: Generation Paradigm} \\
\midrule
\textbf{\textcolor{cobaltdark}{Agents' Room}} \\
\rowcolor{cobaltlight}
\textbf{Multi-agent collaborative narrative generation} baseline. Decomposes the writing process into specialized planning and writing agents communicating via a shared scratchpad to maintain long-term coherence across the plot. \\
\midrule
\textbf{\textcolor{cobaltdark}{WizardLM-StoryTelling-30B}} \\
\rowcolor{cobaltlight}
\textbf{Monolithic instruction-tuning} baseline. Relies on the Evol-Instruct methodology to embed narrative constraints directly into model weights rather than resolving them through external orchestration or preference alignment. \\
\bottomrule

\end{longtable}

\section{Quality-Stratified Analysis: Per-Stratum Results}
\label{sec:appendix-quality-strata}

This appendix provides the detailed statistics for the quality-stratified analysis summarized in Section~\ref{sec:quality-stratified}. For each quality stratum, Table~\ref{tab:quality_strata_jury} reports mean jury scores for the original and PlotTwist-generated plots, bootstrap $95\%$ confidence intervals over paired differences, paired effect sizes (Cohen's $d_z$), the probability that a generated plot outperforms its paired original, and Welch's $t$-tests for secondary statistical validation.

\textbf{Excellent Category (IMDb $>8$).}
Generated plots exhibit only modest improvements over already strong originals. Character development ($+0.50$), narrative coherence ($+0.40$), and tone consistency ($+0.33$) show the clearest gains, whereas emotional turning points improve only marginally ($+0.32$) and pacing remains unchanged ($-0.02$). These results indicate that PlotTwist primarily performs conservative refinement when the original narrative quality is already high.

\textbf{Good Category ($7<$ IMDb $\leq8$).}
Generated plots improve consistently across all NQDs, with the largest gains in character development ($+0.86$), narrative coherence ($+0.85$), and tone consistency ($+0.64$). All dimensions exhibit large effect sizes and statistically significant improvements, indicating systematic enhancement of narratives with solid foundations but remaining structural limitations.

\textbf{Mid Category ($6<$ IMDb $\leq7$).}
Mid-quality narratives benefit substantially from PlotTwist. Narrative coherence ($+1.34$), tone consistency ($+1.15$), and character development ($+1.15$) show the largest improvements, accompanied by uniformly large effect sizes and high dominance probabilities. This quality range represents the regime where PlotTwist performs the most effective narrative restructuring.

\textbf{Low Category (IMDb $\leq6$).}
Generated plots substantially outperform the originals across every NQD. The largest gains occur in narrative coherence ($+1.82$) and tone consistency ($+1.40$), followed by character development ($+1.25$). Generated plots outperform their paired originals on at least $97.5\%$ of films across every dimension, indicating near-complete narrative regeneration for weak source plots.
\begin{table*}[t]
\centering
\small
\setlength{\tabcolsep}{4pt}
\resizebox{\textwidth}{!}{%
\begin{tabular}{ll|cc|ccccc}
\toprule
\textbf{Stratum}
& \textbf{NQD}
& \textbf{Orig.}
& \textbf{Gen.}
& \textbf{$\Delta$}
& \textbf{95\% CI}
& \textbf{$d_z$}
& \textbf{$P(\mathrm{gen}>\mathrm{orig})$}
& \textbf{Welch $p$} \\
\midrule

Excellent
& Narrative coherence
& 8.24 & 8.64 & $+0.40$ & $[0.13,\,0.72]$
& 0.41 & 0.70 & 0.012 \\

($n=40$)
& Emotional turning points
& 8.37 & 8.69 & $+0.32$ & $[0.02,\,0.72]$
& 0.28 & 0.55 & 0.088 \\

& Character development
& 7.72 & 8.22 & $+0.50$ & $[0.20,\,0.88]$
& 0.44 & 0.75 & 0.021 \\

& Pacing
& 8.18 & 8.16 & $-0.02$ & $[-0.21,\,0.19]$
& $-0.04$ & 0.38 & 0.828 \\

& Tone consistency
& 8.18 & 8.51 & $+0.33$ & $[0.17,\,0.50]$
& 0.62 & 0.75 & $4.47\times10^{-4}$ \\

& \textbf{Overall}
& \textbf{8.14} & \textbf{8.45} & $\mathbf{+0.31}$
& $\mathbf{[0.09,\,0.57]}$
& \textbf{0.38} & \textbf{0.65} & \textbf{0.023} \\

\midrule

Good
& Narrative coherence
& 7.70 & 8.55 & $+0.85$ & $[0.64,\,1.08]$
& 1.18 & 0.85 & $3.28\times10^{-8}$ \\

($n=40$)
& Emotional turning points
& 8.19 & 8.65 & $+0.46$ & $[0.34,\,0.59]$
& 1.11 & 0.85 & $2.88\times10^{-7}$ \\

& Character development
& 7.36 & 8.22 & $+0.86$ & $[0.66,\,1.08]$
& 1.24 & 0.88 & $1.05\times10^{-8}$ \\

& Pacing
& 7.68 & 8.12 & $+0.43$ & $[0.28,\,0.59]$
& 0.87 & 0.80 & $1.50\times10^{-6}$ \\

& Tone consistency
& 7.83 & 8.46 & $+0.64$ & $[0.47,\,0.81]$
& 1.16 & 0.95 & $4.15\times10^{-10}$ \\

& \textbf{Overall}
& \textbf{7.75} & \textbf{8.40} & $\mathbf{+0.65}$
& $\mathbf{[0.51,\,0.79]}$
& \textbf{1.40} & \textbf{0.93}
& $\mathbf{2.91\times10^{-11}}$ \\

\midrule

Mid
& Narrative coherence
& 7.20 & 8.54 & $+1.34$ & $[1.08,\,1.63]$
& 1.52 & 0.98 & $5.39\times10^{-10}$ \\

($n=40$)
& Emotional turning points
& 8.08 & 8.65 & $+0.57$ & $[0.43,\,0.73]$
& 1.17 & 0.95 & $1.92\times10^{-7}$ \\

& Character development
& 7.03 & 8.18 & $+1.15$ & $[0.91,\,1.39]$
& 1.44 & 0.93 & $1.80\times10^{-8}$ \\

& Pacing
& 7.44 & 8.10 & $+0.67$ & $[0.48,\,0.89]$
& 1.01 & 0.85 & $6.48\times10^{-8}$ \\

& Tone consistency
& 7.25 & 8.40 & $+1.15$ & $[0.96,\,1.35]$
& 1.82 & 1.00 & $1.59\times10^{-13}$ \\

& \textbf{Overall}
& \textbf{7.40} & \textbf{8.37} & $\mathbf{+0.98}$
& $\mathbf{[0.81,\,1.17]}$
& \textbf{1.68} & \textbf{0.98}
& $\mathbf{2.11\times10^{-11}}$ \\

\midrule

Low
& Narrative coherence
& 6.67 & 8.50 & $+1.82$ & $[1.52,\,2.16]$
& 1.76 & 0.98 & $1.04\times10^{-13}$ \\

($n=40$)
& Emotional turning points
& 7.68 & 8.58 & $+0.90$ & $[0.69,\,1.13]$
& 1.26 & 0.98 & $3.39\times10^{-9}$ \\

& Character development
& 6.59 & 7.84 & $+1.25$ & $[1.00,\,1.53]$
& 1.46 & 0.98 & $2.04\times10^{-9}$ \\

& Pacing
& 7.15 & 8.15 & $+1.00$ & $[0.83,\,1.20]$
& 1.63 & 1.00 & $1.62\times10^{-12}$ \\

& Tone consistency
& 6.94 & 8.35 & $+1.40$ & $[1.19,\,1.61]$
& 2.05 & 0.98 & $7.28\times10^{-16}$ \\

& \textbf{Overall}
& \textbf{7.01} & \textbf{8.28} & $\mathbf{+1.28}$
& $\mathbf{[1.09,\,1.48]}$
& \textbf{2.01} & \textbf{1.00}
& $\mathbf{7.95\times10^{-15}}$ \\

\bottomrule
\end{tabular}%
}
\caption{
Per-stratum jury statistics for original and PlotTwist-generated plots.
$\Delta$ denotes the mean paired difference between generated and original
plots, with bootstrap 95\% confidence intervals.
The statistic $d_z$ is the paired Cohen's effect size, and
$P(\mathrm{gen}>\mathrm{orig})$ is the proportion of films for which the
generated plot receives a higher score than its paired original.
}
\label{tab:quality_strata_jury}
\end{table*}

\textbf{Summary.}
Overall improvements increase monotonically across quality strata: $+0.31$ (Excellent), $+0.65$ (Good), $+0.98$ (Mid), and $+1.28$ (Low), demonstrating that PlotTwist adapts the magnitude of its intervention to the underlying quality of the source narrative rather than uniformly inflating evaluation scores.

\phantomsection
\section{\texorpdfstring{Jury Evaluation: Protocol, Win Rates, and Reliability}{Jury Evaluation: Protocol, Win Rates, and Reliability}}
\label{sec:appendix-jury-details}

\textbf{Composition and serving.} Table~\ref{tab:jury_composition} lists the jury. No juror shares a model family with another juror, with the plot generator, or with the single-judge Qwen3-32B evaluator used during development (Section~\ref{agentic-evaluation}). All five are open-weight models, served locally at pinned revisions and scored at temperature $0.1$. Each run records every juror's model revision, quantization, serving engine, and decoding parameters in a manifest, so all scores can be regenerated.

\begin{table}[ht]
\centering
\small
\begin{tabular}{l|ll}
\toprule[1.1pt]
\textbf{Juror} & \textbf{Family} & \textbf{Precision} \\
\midrule[1.1pt]
GPT-OSS-120B & OpenAI (open-weight) & MXFP4 \\
Gemma-4-31B & Google & BF16 \\
Mistral-Medium-3.5-128B & Mistral & Q6\_K \\
Nemotron-3-Super-120B-A12B & NVIDIA & FP8 \\
GLM-4.6V & Zhipu & FP8 \\
\bottomrule[1.1pt]
\end{tabular}
\caption{Jury composition. The five jurors span five model families, all disjoint from the Qwen family of the plot generator and of the single-judge development evaluator.}
\label{tab:jury_composition}
\end{table}

\textbf{Grounded output contract.} Each juror scores one NQD per call, using the corresponding rubric of Appendix~\ref{sec:appendix-agentic-evaluation-prompts}. The response is a structured JSON object giving, for each of the ten criteria, a score in $[0,1]$, a one-sentence rationale, and the verbatim evidence span required by Section~\ref{sec:jury-evaluation}; the token-containment threshold of $0.92$ tolerates minor quoting drift, and unverifiable citations are logged and reported rather than silently discarded. When a juror's stated total disagrees with the sum of its criterion scores by more than one point, we use the sum, so reported totals always reflect the cited evidence. Enforcing grounding at any verification threshold leaves the ranking of the twelve systems essentially unchanged (Spearman $\geq 0.99$ against the full panel) and moves the two closest win rates by at most $0.04$.

\textbf{Per-judge win rates.} Table~\ref{tab:jury_winrate} in Section~\ref{sec:jury-evaluation} reports the panel-level win rates; here we break them down by judge. The GPT-4.1 comparison is unanimous at the level of individual jurors: each of the five prefers PlotTwist on a majority of premises, with per-judge win rates between $0.65$ and $0.90$, and removing any single juror leaves the panel estimate between $0.85$ and $0.91$, with every bootstrap interval excluding $0.5$. The Claude Sonnet~4 comparison is contested: four jurors individually favor PlotTwist (win rates between $0.59$ and $0.63$), while Gemma-4-31B, the most conservative scorer on the panel, favors Claude ($0.34$). Removing any single juror keeps the estimate between $0.49$ and $0.69$, and no analysis choice we examined turns the comparison into a statistically significant loss.

\textbf{Agreement by dimension.} Table~\ref{tab:jury_agreement} breaks inter-juror agreement down by dimension. Krippendorff's $\alpha$ is lowest on the most subjective dimensions (emotional turning points, tone consistency), while the skew-robust Gwet's AC2 is high on every dimension, the pattern produced by judges who agree on order but differ in leniency. Recentering each judge's scores raises pooled $\alpha$ from $0.413$ to $0.641$, confirming that much of the nominal disagreement is a per-judge level offset. High agreement should not be read as five independent confirmations, however: the Kish effective judge count is $1.35$ of $5$ on raw scores, and $1.81$ after removing system and premise effects \cite{kish1965survey,kohli2026judgeseffectivevotescorrelated}. The per-judge and leave-one-judge-out analyses in the previous paragraph are therefore the ones that carry evidential weight.

\begin{table}[ht]
\centering
\small
\begin{tabular}{l|cc}
\toprule[1.1pt]
\textbf{NQD} & \textbf{Krippendorff's $\alpha$} & \textbf{Gwet's AC2} \\
\midrule[1.1pt]
Narrative coherence & 0.504 & 0.889 \\
Emotional turning points & 0.280 & 0.827 \\
Character development & 0.405 & 0.863 \\
Pacing & 0.476 & 0.932 \\
Tone consistency & 0.366 & 0.911 \\
\midrule
\rowcolor{gray!15}
\textbf{Pooled} & \textbf{0.413} & \textbf{0.887} \\
\bottomrule[1.1pt]
\end{tabular}
\caption{Inter-juror agreement per NQD (interval-level Krippendorff's $\alpha$; ordinal-weighted Gwet's AC2). The pooled AC2 $95\%$ CI is $[0.880, 0.893]$.}
\label{tab:jury_agreement}
\end{table}

\textbf{Same-family bias audit.} To quantify the circularity that motivated the jury, a separate diagnostic run adds an in-family Qwen judge to the panel. Controlling for system identity and premise difficulty, judges score generators from their own family $0.82$ points higher (premise-clustered bootstrap $95\%$ CI $[0.80, 0.84]$), so same-family judging does inflate scores. PlotTwist does not benefit from this effect: its leniency-corrected same-family inflation is negative ($-0.47$), and removing all same-family cells from the panel leaves the ranking of the twelve systems unchanged (Spearman $1.0$).

\textbf{Specification-curve robustness.} We recompute the two closest comparisons under every defensible combination of analysis choices \cite{simonsohn2020specification}: juror subset (all five, or each juror removed), aggregation across jurors (mean, median, trimmed mean), score basis (overall mean or per-NQD majority vote), tie handling (ties as $0.5$, dropped, or as losses), grounding enforcement, and length control \cite{dubois2024length}, for a total of $113$ specifications per comparison. Against GPT-4.1, the win rate remains within $[0.80, 0.92]$ and every specification's $95\%$ interval excludes $0.5$. Against Claude Sonnet~4, the median win rate is $0.556$, $94.7\%$ of specifications lie at or above $0.5$, and none yields a statistically significant loss.

\textbf{Deliberation and final judge.} Deliberation targets only the cells where the panel genuinely disagrees, defined as a spread of at least two points between scores. On each such cell, jurors are shown the anonymized, order-shuffled rationales and cited spans of their peers, with peer scores hidden to prevent vote-matching, and may revise their assessments across two rounds of structured discussion. A held-out final judge, Llama-3.3-70B, drawn from a sixth model family disjoint from both the panel and the generator, then reads the plot, the arguments, and the revised assessments, and issues its own verdict on each contested cell rather than counting votes \cite{chan2024chateval}; panel scores are retained on uncontested cells. Deliberation itself changes little: inter-juror agreement on overall scores moves only from $\alpha = 0.455$ to $0.458$, two premise-level verdicts flip, and no baseline-level conclusion changes. The final judge reproduces the panel's conclusions, with win rates of $0.89$ against GPT-4.1, $0.92$ against Gemini 2.0 Flash, $0.74$ against Agents' Room, and $0.57$ against Claude Sonnet~4 ($p = 0.08$, inconclusive as in the main analysis). Although the judge shares a model family with the Llama-3.3-70B baseline, it prefers PlotTwist against that baseline on all but one premise, so its verdicts are not driven by family loyalty. Because deliberation neither materially raises agreement nor changes any conclusion, the consolidated post-deliberation scores and the independent round-one panel are effectively interchangeable.

\section{Computational Resources}
\addcontentsline{toc}{section}{Computational Resources}
\label{sec:appendix-compute}

\noindent All open-weight models used in this work were downloaded from the \textbf{Hugging Face Model Hub}\footnote{\url{https://huggingface.co}} and executed on a dedicated compute cluster consisting of \textbf{4 $\times$ NVIDIA L40S GPUs}. The L40S is a high-performance data center GPU featuring 48\,GB of GDDR6 memory per card, providing a total of \textbf{192\,GB of aggregate GPU memory}, which was sufficient to accommodate the quantized and sparse model configurations employed throughout the framework.

\vspace{6pt}
\noindent Closed-source frontier models were \textbf{not} run on local infrastructure and were instead accessed exclusively via their respective commercial APIs.

\vspace{10pt}

\begin{longtable}{p{0.32\linewidth} p{0.18\linewidth} p{0.40\linewidth}}
\toprule
\textbf{Model} & \textbf{Execution} & \textbf{Usage in PlotTwist} \\
\midrule

\multicolumn{3}{l}{\textit{\textbf{PlotTwist Core Components --- Local (4 $\times$ L40S)}}} \\
\midrule
\rowcolor{cobaltlight}
Qwen-3-30B-A3B (MoE)   & Local & DPO fine-tuning and inference for PlotTwist Plot Generator \\
Qwen-3-32B (4-bit)     & Local & SFT training of Aspect Rating Reward Model \\
\rowcolor{cobaltlight}
Qwen-3-32B (16-bit)    & Local & Inference for Agentic Evaluation module \\

\midrule
\multicolumn{3}{l}{\textit{\textbf{Ensemble Rating Models --- Local (4 $\times$ L40S)}}} \\
\midrule
\rowcolor{cobaltlight}
Qwen-2.5-7B            & Local & Positive--negative aspect rating ensemble \\
Llama-3.1-8B           & Local & Positive--negative aspect rating ensemble \\
\rowcolor{cobaltlight}
DeepSeek-14B           & Local & Positive--negative aspect rating ensemble \\
Gemma-27B              & Local & Positive--negative aspect rating ensemble; premise generation \\
\rowcolor{cobaltlight}
Llama-3.3-70B & Local & Positive--negative aspect rating ensemble \\

\midrule
\multicolumn{3}{l}{\textit{\textbf{Open-Weight Baselines --- Local (4 $\times$ L40S)}}} \\
\midrule
\rowcolor{cobaltlight}
Llama-3.3-70B & Local & Model scale baseline \\
Qwen3-32B (dense)      & Local & Architectural design baseline \\
\rowcolor{cobaltlight}
Qwen2.5-7B-Instruct    & Local & Scale ablation baseline \\
Qwen2.5-14B-Instruct   & Local & Scale ablation baseline \\
\rowcolor{cobaltlight}
DeepSeek-R1 14B        & Local & Architectural design baseline \\
Phi-4 Mini Instruct    & Local & Architectural design baseline \\
\rowcolor{cobaltlight}
Mistral Small 2501 24B & Local & Generation paradigm baseline \\
WizardLM-30B           & Local & Generation paradigm baseline \\

\midrule
\multicolumn{3}{l}{\textit{\textbf{Frontier Models --- API Access Only}}} \\
\midrule
\rowcolor{cobaltlight}
GPT-4.1                & OpenAI API      & DPO candidate plot generation; model scale baseline \\
Claude Sonnet 4        & Anthropic API   & DPO candidate plot generation; model scale baseline \\
\rowcolor{cobaltlight}
Gemini 2.0 Flash       & Google API      & DPO candidate plot generation; model scale baseline \\

\bottomrule
\end{longtable}

\vspace{4pt}
\noindent \textit{Note: Agents' Room~\cite{huot2024agents} was evaluated according to its original open-source implementation and run locally on the same infrastructure.}

%% file: references.bib
@article{harper2015movielens,
  title={The movielens datasets: History and context},
  author={Harper, F Maxwell and Konstan, Joseph A},
  journal={Acm transactions on interactive intelligent systems (tiis)},
  volume={5},
  number={4},
  pages={1--19},
  year={2015},
  publisher={Acm New York, NY, USA}
}

@inproceedings{chakrabarty2024art,
  title={Art or artifice? large language models and the false promise of creativity},
  author={Chakrabarty, Tuhin and Laban, Philippe and Agarwal, Divyansh and Muresan, Smaranda and Wu, Chien-Sheng},
  booktitle={Proceedings of the 2024 CHI Conference on Human Factors in Computing Systems},
  pages={1--34},
  year={2024}
}

@article{zheng2023judging,
  title={Judging llm-as-a-judge with mt-bench and chatbot arena},
  author={Zheng, Lianmin and Chiang, Wei-Lin and Sheng, Ying and Zhuang, Siyuan and Wu, Zhanghao and Zhuang, Yonghao and Lin, Zi and Li, Zhuohan and Li, Dacheng and Xing, Eric and others},
  journal={Advances in neural information processing systems},
  volume={36},
  pages={46595--46623},
  year={2023}
}

@article{shazeer2017outrageously,
  title={Outrageously large neural networks: The sparsely-gated mixture-of-experts layer},
  author={Shazeer, Noam and Mirhoseini, Azalia and Maziarz, Krzysztof and Davis, Andy and Le, Quoc and Hinton, Geoffrey and Dean, Jeff},
  journal={arXiv preprint arXiv:1701.06538},
  year={2017}
}

@article{fedus2022switch,
  title={Switch transformers: Scaling to trillion parameter models with simple and efficient sparsity},
  author={Fedus, William and Zoph, Barret and Shazeer, Noam},
  journal={Journal of Machine Learning Research},
  volume={23},
  number={120},
  pages={1--39},
  year={2022}
}

@article{rafailov2023direct,
  title={Direct preference optimization: Your language model is secretly a reward model},
  author={Rafailov, Rafael and Sharma, Archit and Mitchell, Eric and Manning, Christopher D and Ermon, Stefano and Finn, Chelsea},
  journal={Advances in neural information processing systems},
  volume={36},
  pages={53728--53741},
  year={2023}
}

@article{huot2024agents,
  title={Agents' Room: Narrative Generation through Multi-step Collaboration},
  author={Huot, F. and Amplayo, R. K. and Palomaki, J. and Jakobovits, A. S. and Clark, E. and Lapata, M.},
  journal={arXiv preprint arXiv:2410.02603},
  year={2024}
}

@article{qwen2025qwen3,
  title={Qwen3 Technical Report},
  author={Qwen Team},
  journal={arXiv preprint arXiv:2505.09388},
  year={2025}
}

@misc{anthropic2025sonnet4,
  title={Claude 4 Model Card and System Safety},
  author={Anthropic},
  year={2025},
  month={May},
  howpublished={\url{https://www.anthropic.com/news/claude-4}}
}

@misc{openai2025gpt41,
  title={GPT-4.1 System Card},
  author={OpenAI},
  year={2025},
  month={April},
  howpublished={\url{https://openai.com/index/gpt-4-1/}}
}

@misc{mistral2025small3,
  title={Mistral Small 3 (2501) Release},
  author={Mistral AI Team},
  year={2025},
  month={January},
  howpublished={\url{https://mistral.ai/news/mistral-small-3/}}
}

@article{deepseek2025r1,
  title={DeepSeek-R1: Incentivizing Reasoning Capability in LLMs via Reinforcement Learning},
  author={DeepSeek-AI},
  journal={arXiv preprint arXiv:2501.12948},
  year={2025}
}

@article{microsoft2025phi4mini,
  title={Phi-4-Mini Technical Report: Compact yet Powerful Multimodal Language Models via Mixture-of-LoRAs},
  author={Microsoft},
  journal={arXiv preprint arXiv:2503.01743},
  year={2025}
}

@misc{gemini2024flash,
  title={Introducing Gemini 2.0: Our New AI Model for the Agentic Era},
  author={Google DeepMind},
  year={2024},
  month={December},
  howpublished={\url{https://blog.google/technology/google-deepmind/google-gemini-ai-update-december-2024/}}
}

@article{qwen2024qwen25,
  title={Qwen2.5 Technical Report},
  author={Qwen Team},
  journal={arXiv preprint arXiv:2412.15115},
  year={2024}
}

@misc{thebloke2023wizardlm,
  title={WizardLM-Uncensored-SuperCOT-StoryTelling-30B-GPTQ},
  author={TheBloke and Hartford, Eric and Kaiokendev},
  year={2023},
  howpublished={\url{https://huggingface.co/TheBloke/WizardLM-Uncensored-SuperCOT-StoryTelling-30B-GPTQ}}
}

@inproceedings{fan2018hierarchical,
  title     = {Hierarchical Neural Story Generation},
  author    = {Fan, Angela and Lewis, Mike and Dauphin, Yann},
  booktitle = {Proceedings of the 56th Annual Meeting of the Association for Computational Linguistics (ACL)},
  year      = {2018},
  pages     = {889--898}
}

@inproceedings{yao2019plan,
  title={Plan-and-write: Towards better automatic storytelling},
  author={Yao, Lili and Peng, Nanyun and Weischedel, Ralph and Knight, Kevin and Zhao, Dongyan and Yan, Rui},
  booktitle={Proceedings of the AAAI Conference on Artificial Intelligence},
  volume={33},
  pages={7378--7385},
  year={2019}
}

@book{hogan2011affective,
  title={Affective narratology: The emotional structure of stories},
  author={Hogan, Patrick Colm},
  year={2011},
  publisher={U of Nebraska Press}
}

@inproceedings{teleki2025survey,
  title={A Survey on LLMs for Story Generation},
  author={Teleki, Maria and Bengali, Vedangi and Dong, Xiangjue and Janjur, Sai Tejas and Liu, Haoran and Liu, Tian and Wang, Cong and Liu, Ting and Zhang, Yin and Shipman, Frank and others},
  booktitle={Findings of the Association for Computational Linguistics: EMNLP 2025},
  pages={13954--13966},
  year={2025}
}

@article{gurung2025learning,
  title={Learning to reason for long-form story generation},
  author={Gurung, Alexander and Lapata, Mirella},
  journal={arXiv preprint arXiv:2503.22828},
  year={2025}
}

@article{kim2025_llm_creativity_eval,
  title        = {Evaluating Creativity: Can LLMs Be Good Evaluators in Creative Writing Tasks?},
  author       = {Sungeun Kim and Dongsuk Oh},
  journal      = {Applied Sciences},
  year         = {2025},
  volume       = {15},
  number       = {6},
  pages        = {2971},
  doi          = {10.3390/app15062971},
  url          = {https://www.mdpi.com/2076-3417/15/6/2971}
}

@misc{zheng2025_cmlbench,
  title        = {CML-Bench: A Framework for Evaluating and Enhancing LLM-Powered Movie Scripts Generation},
  author       = {Mingzhe Zheng and Dingjie Song and Guanyu Zhou and Jun You and Jiahao Zhan and Xuran Ma and Xinyuan Song and Ser-Nam Lim and Qifeng Chen and Harry Yang},
  year         = {2025},
  eprint       = {2510.06231},
  archivePrefix= {arXiv},
  primaryClass = {cs.CL},
  url          = {https://arxiv.org/abs/2510.06231}
}

@article{ouyang2022training,
  title={Training language models to follow instructions with human feedback},
  author={Ouyang, Long and Wu, Jeffrey and Jiang, Xu and Almeida, Diogo and Wainwright, Carroll and Mishkin, Pamela and Zhang, Chong and Agarwal, Sandhini and Slama, Katarina and Ray, Alex and others},
  journal={Advances in neural information processing systems},
  volume={35},
  pages={27730--27744},
  year={2022}
}

@article{sun2023evaluating,
  title={Evaluating the zero-shot robustness of instruction-tuned language models},
  author={Sun, Jiuding and Shaib, Chantal and Wallace, Byron C},
  journal={arXiv preprint arXiv:2306.11270},
  year={2023}
}

@misc{verga2024poll,
  title={Replacing Judges with Juries: Evaluating {LLM} Generations with a Panel of Diverse Models},
  author={Verga, Pat and Hofst{\"a}tter, Sebastian and Althammer, Sophia and Su, Yixuan and Piktus, Aleksandra and Arkhangorodsky, Arkady and Xu, Minjie and White, Naomi and Lewis, Patrick},
  year={2024},
  eprint={2404.18796},
  archivePrefix={arXiv},
  primaryClass={cs.CL}
}

@inproceedings{panickssery2024self,
  title={{LLM} Evaluators Recognize and Favor Their Own Generations},
  author={Panickssery, Arjun and Bowman, Samuel R. and Feng, Shi},
  booktitle={Advances in Neural Information Processing Systems},
  year={2024}
}

@misc{ye2024justice,
  title={Justice or Prejudice? Quantifying Biases in {LLM}-as-a-Judge},
  author={Ye, Jiayi and Wang, Yanbo and Huang, Yue and Chen, Dongping and Zhang, Qihui and Moniz, Nuno and Gao, Tian and Geyer, Werner and Huang, Chao and Chen, Pin-Yu and Chawla, Nitesh V. and Zhang, Xiangliang},
  year={2024},
  eprint={2410.02736},
  archivePrefix={arXiv},
  primaryClass={cs.CL}
}

@misc{fein2025litbench,
  title={{LitBench}: A Benchmark and Dataset for Reliable Evaluation of Creative Writing},
  author={Fein, Daniel and Russo, Sebastian and Xiang, Violet and Jolly, Kabir and Rafailov, Rafael and Haber, Nick},
  year={2025},
  eprint={2507.00769},
  archivePrefix={arXiv},
  primaryClass={cs.CL}
}

@book{krippendorff2004content,
  title={Content Analysis: An Introduction to Its Methodology},
  author={Krippendorff, Klaus},
  edition={2},
  publisher={Sage Publications},
  address={Thousand Oaks, CA},
  year={2004}
}

@article{gwet2008computing,
  title={Computing inter-rater reliability and its variance in the presence of high agreement},
  author={Gwet, Kilem Li},
  journal={British Journal of Mathematical and Statistical Psychology},
  volume={61},
  number={1},
  pages={29--48},
  year={2008}
}

@article{holm1979simple,
  title={A simple sequentially rejective multiple test procedure},
  author={Holm, Sture},
  journal={Scandinavian Journal of Statistics},
  volume={6},
  number={2},
  pages={65--70},
  year={1979}
}

@article{lakens2017equivalence,
  title={Equivalence tests: A practical primer for t tests, correlations, and meta-analyses},
  author={Lakens, Dani{\"e}l},
  journal={Social Psychological and Personality Science},
  volume={8},
  number={4},
  pages={355--362},
  year={2017}
}

@book{kish1965survey,
  title={Survey Sampling},
  author={Kish, Leslie},
  publisher={John Wiley \& Sons},
  address={New York},
  year={1965}
}

@misc{kohli2026judgeseffectivevotescorrelated,
      title={Nine Judges, Two Effective Votes: Correlated Errors Undermine LLM Evaluation Panels}, 
      author={Guneet Kohli},
      year={2026},
      eprint={2605.29800},
      archivePrefix={arXiv},
      primaryClass={cs.CL},
      url={https://arxiv.org/abs/2605.29800}, 
}

@article{simonsohn2020specification,
  title={Specification curve analysis},
  author={Simonsohn, Uri and Simmons, Joseph P. and Nelson, Leif D.},
  journal={Nature Human Behaviour},
  volume={4},
  number={11},
  pages={1208--1214},
  year={2020}
}

@inproceedings{chan2024chateval,
  title={{ChatEval}: Towards Better {LLM}-based Evaluators through Multi-Agent Debate},
  author={Chan, Chi-Min and Chen, Weize and Su, Yusheng and Yu, Jianxuan and Xue, Wei and Zhang, Shanghang and Fu, Jie and Liu, Zhiyuan},
  booktitle={International Conference on Learning Representations},
  year={2024}
}

@misc{dubois2024length,
  title={Length-Controlled {AlpacaEval}: A Simple Way to Debias Automatic Evaluators},
  author={Dubois, Yann and Galambosi, Bal{\'a}zs and Liang, Percy and Hashimoto, Tatsunori B.},
  year={2024},
  eprint={2404.04475},
  archivePrefix={arXiv},
  primaryClass={cs.LG}
}

@misc{jiang2025hamlet,
  title={{HAMLET}: A Hierarchical and Adaptive Multi-Agent Framework for Live Embodied Theatrics},
  author={Jiang, Shufan and Chen, Sizhou and Chen, Chios and Zhang, Chi and Zhang, Xiao-Lei and Li, Xuelong},
  year={2025},
  eprint={2507.15518},
  archivePrefix={arXiv},
  primaryClass={cs.CL}
}

@misc{nvidia2025nvidianemotron3efficient,
      title={NVIDIA Nemotron 3: Efficient and Open Intelligence}, 
      author={NVIDIA and : and Aaron Blakeman and Aaron Grattafiori and Aarti Basant and Abhibha Gupta and Abhinav Khattar and Adi Renduchintala and Aditya Vavre and Akanksha Shukla and Akhiad Bercovich and Aleksander Ficek and Aleksandr Shaposhnikov and Alex Kondratenko and Alexander Bukharin and Alexandre Milesi and Ali Taghibakhshi and Alisa Liu and Amelia Barton and Ameya Sunil Mahabaleshwarkar and Amir Klein and Amit Zuker and Amnon Geifman and Amy Shen and Anahita Bhiwandiwalla and Andrew Tao and Anjulie Agrusa and Ankur Verma and Ann Guan and Anubhav Mandarwal and Arham Mehta and Ashwath Aithal and Ashwin Poojary and Asif Ahamed and Asit Mishra and Asma Kuriparambil Thekkumpate and Ayush Dattagupta and Banghua Zhu and Bardiya Sadeghi and Barnaby Simkin and Ben Lanir and Benedikt Schifferer and Besmira Nushi and Bilal Kartal and Bita Darvish Rouhani and Boris Ginsburg and Brandon Norick and Brandon Soubasis and Branislav Kisacanin and Brian Yu and Bryan Catanzaro and Carlo del Mundo and Chantal Hwang and Charles Wang and Cheng-Ping Hsieh and Chenghao Zhang and Chenhan Yu and Chetan Mungekar and Chintan Patel and Chris Alexiuk and Christopher Parisien and Collin Neale and Cyril Meurillon and Damon Mosk-Aoyama and Dan Su and Dane Corneil and Daniel Afrimi and Daniel Lo and Daniel Rohrer and Daniel Serebrenik and Daria Gitman and Daria Levy and Darko Stosic and David Mosallanezhad and Deepak Narayanan and Dhruv Nathawani and Dima Rekesh and Dina Yared and Divyanshu Kakwani and Dong Ahn and Duncan Riach and Dusan Stosic and Edgar Minasyan and Edward Lin and Eileen Long and Eileen Peters Long and Elad Segal and Elena Lantz and Ellie Evans and Elliott Ning and Eric Chung and Eric Harper and Eric Tramel and Erick Galinkin and Erik Pounds and Evan Briones and Evelina Bakhturina and Evgeny Tsykunov and Faisal Ladhak and Fay Wang and Fei Jia and Felipe Soares and Feng Chen and Ferenc Galko and Frank Sun and Frankie Siino and Gal Hubara Agam and Ganesh Ajjanagadde and Gantavya Bhatt and Gargi Prasad and George Armstrong and Gerald Shen and Gorkem Batmaz and Grigor Nalbandyan and Haifeng Qian and Harsh Sharma and Hayley Ross and Helen Ngo and Herbert Hum and Herman Sahota and Hexin Wang and Himanshu Soni and Hiren Upadhyay and Huizi Mao and Huy C Nguyen and Huy Q Nguyen and Iain Cunningham and Ido Galil and Ido Shahaf and Igor Gitman and Ilya Loshchilov and Itamar Schen and Itay Levy and Ivan Moshkov and Izik Golan and Izzy Putterman and Jan Kautz and Jane Polak Scowcroft and Jared Casper and Jatin Mitra and Jeffrey Glick and Jenny Chen and Jesse Oliver and Jian Zhang and Jiaqi Zeng and Jie Lou and Jimmy Zhang and Jinhang Choi and Jining Huang and Joey Conway and Joey Guman and John Kamalu and Johnny Greco and Jonathan Cohen and Joseph Jennings and Joyjit Daw and Julien Veron Vialard and Junkeun Yi and Jupinder Parmar and Kai Xu and Kan Zhu and Kari Briski and Katherine Cheung and Katherine Luna and Keith Wyss and Keshav Santhanam and Kevin Shih and Kezhi Kong and Khushi Bhardwaj and Kirthi Shankar and Krishna C. Puvvada and Krzysztof Pawelec and Kumar Anik and Lawrence McAfee and Laya Sleiman and Leon Derczynski and Li Ding and Lizzie Wei and Lucas Liebenwein and Luis Vega and Maanu Grover and Maarten Van Segbroeck and Maer Rodrigues de Melo and Mahdi Nazemi and Makesh Narsimhan Sreedhar and Manoj Kilaru and Maor Ashkenazi and Marc Romeijn and Marcin Chochowski and Mark Cai and Markus Kliegl and Maryam Moosaei and Matt Kulka and Matvei Novikov and Mehrzad Samadi and Melissa Corpuz and Mengru Wang and Meredith Price and Michael Andersch and Michael Boone and Michael Evans and Miguel Martinez and Mikail Khona and Mike Chrzanowski and Minseok Lee and Mohammad Dabbah and Mohammad Shoeybi and Mostofa Patwary and Nabin Mulepati and Najeeb Nabwani and Natalie Hereth and Nave Assaf and Negar Habibi and Neta Zmora and Netanel Haber and Nicola Sessions and Nidhi Bhatia and Nikhil Jukar and Nikki Pope and Nikolai Ludwig and Nima Tajbakhsh and Nir Ailon and Nirmal Juluru and Nishant Sharma and Oleksii Hrinchuk and Oleksii Kuchaiev and Olivier Delalleau and Oluwatobi Olabiyi and Omer Ullman Argov and Omri Puny and Oren Tropp and Ouye Xie and Parth Chadha and Pasha Shamis and Paul Gibbons and Pavlo Molchanov and Pawel Morkisz and Peter Dykas and Peter Jin and Pinky Xu and Piotr Januszewski and Pranav Prashant Thombre and Prasoon Varshney and Pritam Gundecha and Przemek Tredak and Qing Miao and Qiyu Wan and Rabeeh Karimi Mahabadi and Rachit Garg and Ran El-Yaniv and Ran Zilberstein and Rasoul Shafipour and Rich Harang and Rick Izzo and Rima Shahbazyan and Rishabh Garg and Ritika Borkar and Ritu Gala and Riyad Islam and Robert Hesse and Roger Waleffe and Rohit Watve and Roi Koren and Ruoxi Zhang and Russell Hewett and Russell J. Hewett and Ryan Prenger and Ryan Timbrook and Sadegh Mahdavi and Sahil Modi and Samuel Kriman and Sangkug Lim and Sanjay Kariyappa and Sanjeev Satheesh and Saori Kaji and Satish Pasumarthi and Saurav Muralidharan and Sean Narentharen and Sean Narenthiran and Seonmyeong Bak and Sergey Kashirsky and Seth Poulos and Shahar Mor and Shanmugam Ramasamy and Shantanu Acharya and Shaona Ghosh and Sharath Turuvekere Sreenivas and Shelby Thomas and Shiqing Fan and Shreya Gopal and Shrimai Prabhumoye and Shubham Pachori and Shubham Toshniwal and Shuoyang Ding and Siddharth Singh and Simeng Sun and Smita Ithape and Somshubra Majumdar and Soumye Singhal and Stas Sergienko and Stefania Alborghetti and Stephen Ge and Sugam Dipak Devare and Sumeet Kumar Barua and Suseella Panguluri and Suyog Gupta and Sweta Priyadarshi and Syeda Nahida Akter and Tan Bui and Teodor-Dumitru Ene and Terry Kong and Thanh Do and Tijmen Blankevoort and Tim Moon and Tom Balough and Tomer Asida and Tomer Bar Natan and Tomer Ronen and Tugrul Konuk and Twinkle Vashishth and Udi Karpas and Ushnish De and Vahid Noorozi and Vahid Noroozi and Venkat Srinivasan and Venmugil Elango and Victor Cui and Vijay Korthikanti and Vinay Rao and Vitaly Kurin and Vitaly Lavrukhin and Vladimir Anisimov and Wanli Jiang and Wasi Uddin Ahmad and Wei Du and Wei Ping and Wenfei Zhou and Will Jennings and William Zhang and Wojciech Prazuch and Xiaowei Ren and Yashaswi Karnati and Yejin Choi and Yev Meyer and Yi-Fu Wu and Yian Zhang and Yigong Qin and Ying Lin and Yonatan Geifman and Yonggan Fu and Yoshi Subara and Yoshi Suhara and Yubo Gao and Zach Moshe and Zhen Dong and Zhongbo Zhu and Zihan Liu and Zijia Chen and Zijie Yan},
      year={2025},
      eprint={2512.20856},
      archivePrefix={arXiv},
      primaryClass={cs.CL},
      url={https://arxiv.org/abs/2512.20856}, 
}

@misc{gemmateam2026gemma4technicalreport,
      title={Gemma 4 Technical Report}, 
      author={Gemma Team and Sherif El Abd and Vaibhav Aggarwal and Robin Algayres and Alek Andreev and Olivier Bachem and Ian Ballantyne and Cormac Brick and Victor Cărbune and Michelle Casbon and Mayank Chaturvedi and Victor Cotruta and Alice Coucke and Phil Culliton and Robert Dadashi and Lucas Dixon and Mohamed Elhawaty and Utku Evci and Clément Farabet and Johan Ferret and Filippo Galgani and Sertan Girgin and Jean-Bastien Grill and Maarten Grootendorst and Jiaxian Guo and Cassidy Hardin and Yanzhang He and Steven M. Hernandez and Omri Homburger and Léonard Hussenot and Juyeong Ji and Armand Joulin and Aishwarya Kamath and Parnian Kassraie and Olivier Lacombe and Preethi Lahoti and Gaël Liu and Gus Martins and Luciano Martins and Tatiana Matejovicova and Ramona Merhej and Nikola Momchev and Sneha Mondal and Ryan Mullins and Sindhu Raghuram Panyam and Shreya Pathak and Sarah Perrin and André Susano Pinto and Etienne Pot and Angéline Pouget and Alexandre Ramé and Sabela Ramos and Douglas Reid and David Rim and Morgane Rivière and Karsten Roth and Louis Rouillard and Omar Sanseviero and Pier Giuseppe Sessa and Shane Settle and Danila Sinopalnikov and Sara Smoot and Piotr Stanczyk and Andreas Steiner and Lawrence Stewart and Ilya Tolstikhin and Michael Tschannen and Anton Tsitsulin and Nino Vieillard and Renjie Wu and Pingmei Xu and Haichuan Yang and Edouard Yvinec and Li Zhang and Joe Zou and Nicolas Aagnes and Abdelrahman Abdelhamed and Shivani Agrawal and Shubham Agrawal and Ibrahim Alabdulmohsin and Jean Baptiste Alayrac and Uri Alon and Chandramouli Amarnath and Ankesh Anand and Chrysovalantis Anastasiou and Setareh Ariafar and François-Xavier Aubet and Kyriakos Axiotis and Federico Barbero and Joelle Barral and Alexei Bendebury and Urs Bergmann and Stanley Bileschi and Kat Black and Mathieu Blondel and Sebastian Borgeaud and Arthur Bražinskas and Ryan Burnell and Robert Busa-Fekete and Mu Cai and Glenn Cameron and Charlotte Caucheteux and Garima Chadha and Jetha Chan and Aditya Chawla and Blake Jianhang Chen and Jesse Chen and Lin Chen and Xu Chen and Derek Cheng and Tzu-hsiang Chien and Nikolai Chinaev and Yi Chou and Zhaohui Chu and Benjamin Coleman and Pooja Consul and Sam Conway-Rahman and Scott Crowell and Dylan Cutler and Vivek Dani and Samira Daruki and Anil Das and Daniel Deutsch and Nishanth Dikkala and Li Ding and Qiuhan Ding and Shenil Dodhia and Konstantin Donhauser and Tulsee Doshi and Anca Dragan and Alex Druinsky and Sahil Dua and Zoltan Egyed and Danielle Eisenbud and Daniel Eppens and Cindy Fan and Bahare Fatemi and Yassir Fathullah and Vlad Feinberg and Milen Ferev and Takumi Fujimoto and Isaac Galatzer-Levy and João Gante and Simon Geisler and Soham Ghosal and Antonious M. Girgis and Alec Go and Alhaad Gokhale and Alex Grills and Yiming Gu and Pramod Gupta and Guru Guruganesh and Raia Hadsell and Hamza Harkous and Jitendra Harlalka and Demis Hassabis and Anja Hauth and Joe Heyward and Arian Hosseini and Chih-Yang Hsia and I-Hung Hsu and Xiaopeng Huang and Yangsibo Huang and Kevin Hui and Adrian Hutter and Te I and Fotis Iliopoulos and Advait Jain and Ganesh Jawahar and Ziwei Ji and Qilin Jin and Melvin Johnson and Kandarp Joshi and Arun Kandoor and Wang-Cheng Kang and Koray Kavukcuoglu and Mehran Kazemi and Kathleen Kenealy and Amr Khalifa and Phoebe Kirk and Suraj Kothawade and Vitaly Kovalev and Neel Kovelamudi and Adam Kraft and Ravin Kumar and Harish Kuppam and Justin Lannin and Chen-Yu Lee and Seungji Lee and Dmitry Lepikhin and Dongdong Li and Qiujia Li and Valentin Liévin and Ethan Lin and Ziqian Lin and Casper Liu and Tianlin Liu and Tianqi Liu and Xin Liu and Mayank Lunayach and Min Ma and Gagan Madan and Andrii Maksai and Eric Malmi and Michal Matuszak and Daniel McDuff and Gaurav Menghani and Daniil Mirylenka and Karolis Misiunas and Vedant Misra and Andreea Mitran and Kareem Mohamed and Maksim Mukha and Eric Noland and James O'Donnell and Kate Olszewska and Bernett Orlando and Wanqiong Pan and Rina Panigrahy and Unnati Parekh and Chunjong Park and Eric Paskie and Liqian Peng and Bryce Petrini and Slav Petrov and Jonas Pfeiffer and Bilal Piot and Martyna Plomecka and Siim Poder and Octavio Ponce and Arijit Pramanik and David Racz and Anish Rajan and Michelle Ramanovich and Anand Rao and Marvin Ritter and Vitor Rodrigues and Evan Rosen and Mikołaj Rybiński and Noveen Sachdeva and Michaël E. Sander and Rohit Sathyanarayana and Sagar Savla and Samuel Schmidgall and Tal Schuster and Benoit Seguin and Andrew Sellergren and Aliaksei Severyn and Izhak Shafran and Dhruv Shah and Yuan Shangguan and Ashish Shenoy and Pradeep Shenoy and Rakesh Shivanna and Pauline Sho and Lucas Spangher and Wojciech Stokowiec and Tim Strother and Yao Su and Yinghao Sun and Mukund Sundararajan and Andrea Tacchetti and Mor Hazan Taege and Pouya Tafti and Chetan Tekur and Rahul Thapa and Madeleine Traverse and Lenart Treven and Tao Tu and Chien Te Tung and Petar Veličković and Malini Pooni Venkat and Sagar Gubbi Venkatesh and Vidya Venkiteswaran and Francesco Visin and Alex Vitvitskyi and Kiran Vodrahalli and Weiyi Wang and Xin Wang and Tris Warkentin and Jan Wassenberg and John Wieting and Lechao Xiao and Hao Xu and Yuhui Xu and Fuzhao Xue and Arun Yadav and Jun Yan and Antoine Yang and Lin Yang and Ming-Hsuan Yang and Ziyu Ying and Jae Hyeon Yoo and Sajjad Zafar and Fred Zhang and Jiageng Zhang and Jianyi Zhang and Xiaofan Zhang and Chao Zhao and David Zhou and Chen Zou},
      year={2026},
      eprint={2607.02770},
      archivePrefix={arXiv},
      primaryClass={cs.CL},
      url={https://arxiv.org/abs/2607.02770}, 
}

@misc{5team2025glm45agenticreasoningcoding,
      title={GLM-4.5: Agentic, Reasoning, and Coding (ARC) Foundation Models}, 
      author={ 5 Team and Aohan Zeng and Xin Lv and Qinkai Zheng and Zhenyu Hou and Bin Chen and Chengxing Xie and Cunxiang Wang and Da Yin and Hao Zeng and Jiajie Zhang and Kedong Wang and Lucen Zhong and Mingdao Liu and Rui Lu and Shulin Cao and Xiaohan Zhang and Xuancheng Huang and Yao Wei and Yean Cheng and Yifan An and Yilin Niu and Yuanhao Wen and Yushi Bai and Zhengxiao Du and Zihan Wang and Zilin Zhu and Bohan Zhang and Bosi Wen and Bowen Wu and Bowen Xu and Can Huang and Casey Zhao and Changpeng Cai and Chao Yu and Chen Li and Chendi Ge and Chenghua Huang and Chenhui Zhang and Chenxi Xu and Chenzheng Zhu and Chuang Li and Congfeng Yin and Daoyan Lin and Dayong Yang and Dazhi Jiang and Ding Ai and Erle Zhu and Fei Wang and Gengzheng Pan and Guo Wang and Hailong Sun and Haitao Li and Haiyang Li and Haiyi Hu and Hanyu Zhang and Hao Peng and Hao Tai and Haoke Zhang and Haoran Wang and Haoyu Yang and He Liu and He Zhao and Hongwei Liu and Hongxi Yan and Huan Liu and Huilong Chen and Ji Li and Jiajing Zhao and Jiamin Ren and Jian Jiao and Jiani Zhao and Jianyang Yan and Jiaqi Wang and Jiayi Gui and Jiayue Zhao and Jie Liu and Jijie Li and Jing Li and Jing Lu and Jingsen Wang and Jingwei Yuan and Jingxuan Li and Jingzhao Du and Jinhua Du and Jinxin Liu and Junkai Zhi and Junli Gao and Ke Wang and Lekang Yang and Liang Xu and Lin Fan and Lindong Wu and Lintao Ding and Lu Wang and Man Zhang and Minghao Li and Minghuan Xu and Mingming Zhao and Mingshu Zhai and Pengfan Du and Qian Dong and Shangde Lei and Shangqing Tu and Shangtong Yang and Shaoyou Lu and Shijie Li and Shuang Li and Shuang-Li and Shuxun Yang and Sibo Yi and Tianshu Yu and Wei Tian and Weihan Wang and Wenbo Yu and Weng Lam Tam and Wenjie Liang and Wentao Liu and Xiao Wang and Xiaohan Jia and Xiaotao Gu and Xiaoying Ling and Xin Wang and Xing Fan and Xingru Pan and Xinyuan Zhang and Xinze Zhang and Xiuqing Fu and Xunkai Zhang and Yabo Xu and Yandong Wu and Yida Lu and Yidong Wang and Yilin Zhou and Yiming Pan and Ying Zhang and Yingli Wang and Yingru Li and Yinpei Su and Yipeng Geng and Yitong Zhu and Yongkun Yang and Yuhang Li and Yuhao Wu and Yujiang Li and Yunan Liu and Yunqing Wang and Yuntao Li and Yuxuan Zhang and Zezhen Liu and Zhen Yang and Zhengda Zhou and Zhongpei Qiao and Zhuoer Feng and Zhuorui Liu and Zichen Zhang and Zihan Wang and Zijun Yao and Zikang Wang and Ziqiang Liu and Ziwei Chai and Zixuan Li and Zuodong Zhao and Wenguang Chen and Jidong Zhai and Bin Xu and Minlie Huang and Hongning Wang and Juanzi Li and Yuxiao Dong and Jie Tang},
      year={2025},
      eprint={2508.06471},
      archivePrefix={arXiv},
      primaryClass={cs.CL},
      url={https://arxiv.org/abs/2508.06471}, 
}

@misc{openai2025gptoss120bgptoss20bmodel,
      title={gpt-oss-120b \& gpt-oss-20b Model Card},
      author={OpenAI and : and Sandhini Agarwal and Lama Ahmad and Jason Ai and Sam Altman and Andy Applebaum and Edwin Arbus and Rahul K. Arora and Yu Bai and Bowen Baker and Haiming Bao and Boaz Barak and Ally Bennett and Tyler Bertao and Nivedita Brett and Eugene Brevdo and Greg Brockman and Sebastien Bubeck and Che Chang and Kai Chen and Mark Chen and Enoch Cheung and Aidan Clark and Dan Cook and Marat Dukhan and Casey Dvorak and Kevin Fives and Vlad Fomenko and Timur Garipov and Kristian Georgiev and Mia Glaese and Tarun Gogineni and Adam Goucher and Lukas Gross and Katia Gil Guzman and John Hallman and Jackie Hehir and Johannes Heidecke and Alec Helyar and Haitang Hu and Romain Huet and Jacob Huh and Saachi Jain and Zach Johnson and Chris Koch and Irina Kofman and Dominik Kundel and Jason Kwon and Volodymyr Kyrylov and Elaine Ya Le and Guillaume Leclerc and James Park Lennon and Scott Lessans and Mario Lezcano-Casado and Yuanzhi Li and Zhuohan Li and Ji Lin and Jordan Liss and Lily and Liu and Jiancheng Liu and Kevin Lu and Chris Lu and Zoran Martinovic and Lindsay McCallum and Josh McGrath and Scott McKinney and Aidan McLaughlin and Song Mei and Steve Mostovoy and Tong Mu and Gideon Myles and Alexander Neitz and Alex Nichol and Jakub Pachocki and Alex Paino and Dana Palmie and Ashley Pantuliano and Giambattista Parascandolo and Jongsoo Park and Leher Pathak and Carolina Paz and Ludovic Peran and Dmitry Pimenov and Michelle Pokrass and Elizabeth Proehl and Huida Qiu and Gaby Raila and Filippo Raso and Hongyu Ren and Kimmy Richardson and David Robinson and Bob Rotsted and Hadi Salman and Suvansh Sanjeev and Max Schwarzer and D. Sculley and Harshit Sikchi and Kendal Simon and Karan Singhal and Yang Song and Dane Stuckey and Zhiqing Sun and Philippe Tillet and Sam Toizer and Foivos Tsimpourlas and Nikhil Vyas and Eric Wallace and Xin Wang and Miles Wang and Olivia Watkins and Kevin Weil and Amy Wendling and Kevin Whinnery and Cedric Whitney and Hannah Wong and Lin Yang and Yu Yang and Michihiro Yasunaga and Kristen Ying and Wojciech Zaremba and Wenting Zhan and Cyril Zhang and Brian Zhang and Eddie Zhang and Shengjia Zhao},
      year={2025},
      eprint={2508.10925},
      archivePrefix={arXiv},
      primaryClass={cs.CL},
      url={https://arxiv.org/abs/2508.10925}, 
}
